\documentclass[10pt,journal,compsoc]{IEEEtran}
\usepackage{cite}
\usepackage{amsmath,amssymb,amsfonts}
\usepackage{xcolor}
\usepackage{color}
\usepackage{times}
\usepackage{epsfig}
\usepackage{graphicx}
\graphicspath{{images/}}
\usepackage{verbatim}
\usepackage{subfig}
\usepackage{array}
\usepackage{float}
\usepackage{multirow}
\usepackage{multicol}
\usepackage{siunitx}
\usepackage{hyperref}
\usepackage[english]{babel}
\usepackage[utf8]{inputenc}
\usepackage{algorithm}
\usepackage[noend]{algpseudocode}
\usepackage{booktabs}
\usepackage{colortbl}
\usepackage{enumitem}
\usepackage{listings}
\usepackage{tablefootnote}
\usepackage{cite}
\usepackage{wrapfig, blindtext}
\usepackage{hyperref}
\hypersetup{colorlinks=true, linkcolor=blue, filecolor=magenta, urlcolor=magenta}

\usepackage[flushleft]{threeparttable}
\definecolor{mygray}{gray}{.9}
\definecolor{myblue}{rgb}{0.1,0.1,1}
\usepackage[full,warn]{textcomp}

\ifCLASSINFOpdf
\else
\fi

\begin{document}
%
\title{Automatic Gaze Analysis: A Survey of Deep Learning based Approaches}
%
%
%
%

\author{Shreya~Ghosh, Abhinav~Dhall, Munawar~Hayat, Jarrod~Knibbe, Qiang Ji
\IEEEcompsocitemizethanks{
\IEEEcompsocthanksitem S. Ghosh and M. Hayat are with Monash University. (E-mail: \{shreya.ghosh, munawar.hayat\}@monash.edu).
\IEEEcompsocthanksitem A. Dhall is with Monash University and Indian Institute of Technology Ropar, India. (E-mail: abhinav.dhall@monash.edu).
\IEEEcompsocthanksitem J. Knibbe is with the University of Melbourne. (E-mail: jarrod.knibbe@unimelb.edu.au).
\IEEEcompsocthanksitem Q. Ji is with the Rensselaer Polytechnic Institute (E-mail: jiq@rpi.edu).
}}

%
%

\markboth{
}
{Ghosh \MakeLowercase{\textit{et al.}}: Automatic Gaze Analysis: A Survey of Deep Learning based Approaches}
%



\IEEEtitleabstractindextext{%
\begin{abstract}
Eye gaze analysis is an important research problem in the field of Computer Vision and Human-Computer Interaction. Even with notable progress in the last 10 years, automatic gaze analysis still remains challenging due to the uniqueness of eye appearance, eye-head interplay, occlusion, image quality, and illumination conditions. There are several open questions, including what are the important cues to interpret gaze direction in an unconstrained environment without prior knowledge and how to encode them in real-time. We review the progress across a range of gaze analysis tasks and applications to elucidate these fundamental questions, identify effective methods in gaze analysis, and provide possible future directions. We analyze recent gaze estimation and segmentation methods, especially in the unsupervised and weakly supervised domain, based on their advantages and reported evaluation metrics. Our analysis shows that the development of a robust and generic gaze analysis method still needs to address real-world challenges such as unconstrained setup and learning with less supervision. We conclude by discussing future research directions for designing a real-world gaze analysis system that can propagate to other domains including Computer Vision, Augmented Reality (AR), Virtual Reality (VR), and Human Computer Interaction (HCI). Project Page: \href{https://github.com/i-am-shreya/EyeGazeSurvey}{https://github.com/i-am-shreya/EyeGazeSurvey}

\end{abstract}

}

\maketitle

\IEEEdisplaynontitleabstractindextext

%
\IEEEpeerreviewmaketitle

\IEEEraisesectionheading{\section{Introduction}\label{sec:introduction}}

%
%
%
%
\IEEEPARstart{H}{umans} perceive their environment through voluntary or involuntary eye movement to receive, fixate and track visual stimuli, or in response to an auditory, or cognitive stimulus. The eye movements therefore can provide insights into our visual attention~\cite{liu2011visual} and cognition (emotions, beliefs and desires)~\cite{frischen2007gaze}. Furthermore, we rely on these insights extensively in day-to-day communication and social interaction.


Automatic gaze analysis develops techniques to estimate the position of target objects by observing the eyes' movement. However, accurate gaze analysis is a complex problem. An accurate method should be able to disentangle gaze, while being resilient to a broad array of challenges, including: eye-head interplay, illumination variations, eye registration errors, occlusions, and identity bias. Furthermore, research~\cite{purves2015perception} has shown how human gaze follows an arbitrary trajectory during eye movements which poses further challenge in gaze estimation. 

\begin{figure*}[t]
  \centering
      \includegraphics[width=0.99 \linewidth]{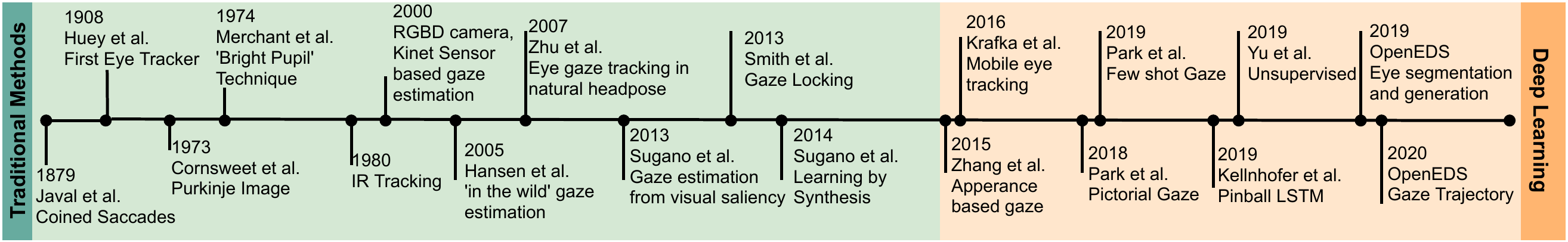}
      \put(-475,12.5){\scriptsize{\cite{javal1878essai}}}
      \put(-464,65){\scriptsize{\cite{huey1908psychology}}}
      \put(-405,65){\scriptsize{\cite{merchant1974remote}}}
      \put(-415.7,22.5){\scriptsize{\cite{cornsweet1973accurate}}}
      \put(-328,22){\scriptsize{\cite{hansen2005eye}}}
      \put(-316,62){\scriptsize{\cite{zhu2007novel}}}
      \put(-273,26){\scriptsize{\cite{sugano2012appearance}}}
      \put(-240,58){\scriptsize{\cite{smith2013gaze}}}
      \put(-220,26){\scriptsize{\cite{sugano2014learning}}}
      \put(-173,23){\scriptsize{\cite{zhang2015appearance}}}
      \put(-132,21){\scriptsize{\cite{park2018deep}}}
      \put(-168,63){\scriptsize{\cite{krafka2016eye}}}
      \put(-125,60){\scriptsize{\cite{park2019few}}}
      \put(-83.4,60){\scriptsize{\cite{yu2019unsupervised}}}
      \put(-40,60){\scriptsize{\cite{garbin2019openeds}}}
      \put(-75,20){\scriptsize{\cite{kellnhofer2019gaze360}}}
      \put(-61.6,5.8){\scriptsize{\cite{palmero2020openeds2020}}}
\caption
{A brief chronology of seminal gaze analysis works. The very first gaze pattern modelling dates back to the work of Javal et al. in 1879~\cite{javal1878essai}. One of the first deep learning driven appearance based gaze estimation models {was proposed in} $2015$~\cite{zhang2015appearance}. 
}

\label{fig:chronology}
\vspace{-5mm}
\end{figure*}
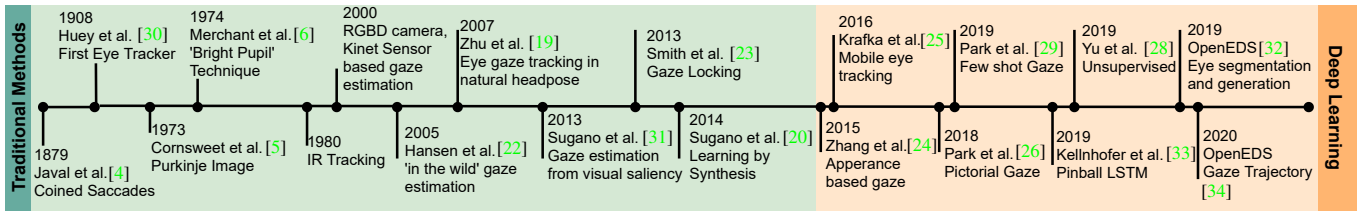

Research in gaze analysis mainly involves coarse or fine-grained gaze estimation. There are three aspects of gaze analysis: registration, representation, and recognition. The first step, \textit{registration}, involves the detection of the eyes (or eye-related key points or sometimes even just the face). In the second step, \textit{representation}, the detected eye is projected to a meaningful feature space. In the final stage, \textit{recognition}, the corresponding gaze direction or gaze location is predicted based on the features from stage 2.
Research interest in automatic gaze analysis spans in several disciplines. One of the earliest explorations of gaze analysis was conducted in 1879, when Javal et al.~\cite{javal1878essai} first studied, and coined the term, \textit{saccades}. The broader interest in gaze analysis, however, developed with the advent of eye tracking technologies (initially in 1908, before gaining momentum in the late 70s, with systems such as `Purkinje Image'\cite{cornsweet1973accurate}, `Bright Pupil'~\cite{merchant1974remote}). Automated gaze analysis then gained traction in computer vision-related assistive technology~\cite{borgestig2016eye,corno2002cost}, which then propagated through HCI~\cite{joseph2020potential,pi2020dynamic,chen2019unsupervised}, consumer behavior analysis~\cite{wedel2017review}, AR and VR~\cite{patney2016perceptually,azuma1997survey}, egocentric vision~\cite{ragusa2020ego}, biometric systems~\cite{jain2006biometrics} and other domains~\cite{eckstein2017beyond,miller2011persistence}. A brief chronology of the seminal gaze analysis methods with important milestones is presented in Fig.~\ref{fig:chronology}. 
The increased reliance on gaze tracking technologies, however, came with its own challenges, namely, the cost of such devices and the requirement for specific controlled settings. To overcome their limitations and handle generic unconstrained settings~\cite{zhu2007novel,sugano2014learning}, most traditional gaze analysis models rely on handcrafted low-level features (e.g., color~\cite{hansen2009eye}, shape~\cite{hansen2005eye,hansen2009eye} and appearance~\cite{smith2013gaze}) and certain geometrical heuristics~\cite{sugano2014learning}. Since 2015, the approach to gaze analysis has changed, turning to the deep learning~\cite{zhang2015appearance,krafka2016eye,park2018deep}, similar to other computer vision tasks. With the deep learning based models and the availability of the large training datasets, the challenges associated with the variation in lighting, camera setup, eye-head interplay, etc, are reduced greatly over the past few years. Although, these performance enhancements have come with the requirement of large scale annotated data, which is expensive to acquire. As such, more recently, deep learning with limited annotation has gained increasing popularity ~\cite{dubey2019unsupervised,yu2019unsupervised,park2019few}.

This paper surveys different gaze analysis methods by isolating their fundamental components, and discusses how each component addresses the aforementioned challenges in gaze analysis. The paper discusses new trends and developments in the field of computer vision and the AR/VR domain, from the perspective of gaze analysis.
We cover recent gaze analysis techniques in the un-, self-, and weakly-supervised domain, along with validation protocols with evaluation metrics tailored for gaze analysis. We also discuss various data capturing devices, including: RGB/IR camera, tablet/laptop's camera, ladybug camera and other gaze trackers (including video-oculography~\cite{hansen2009eye}) are also discussed. 

Due to the rapid progress in the computer vision field (Refer Fig.~\ref{fig:chronology}), it is increasingly useful to get thorough guidance via exhaustive survey/review articles. In 2010 and 2013, Hansen et al.~\cite{hansen2009eye} and Chennamma et al.~\cite{chennamma2013survey} reviewed the state-of-the-art eye detection and gaze tracking techniques. These reviews provide a holistic view of hardware, user interface, eye detection, and gaze mapping techniques. Since these reviews were before the deep learning era, they contain the relevant features leveraged from handcrafted techniques. Afterwards in 2016, Jing et al.~\cite{jing2016survey} reviewed methods for 2-D and 3-D gaze estimation methods. In 2017, Kar et al.~\cite{kar2017review} provided insights into the issues related to algorithms, system configurations, and user conditions. In 2020, a more comprehensive and detailed study of deep learning based gaze estimation methods is presented by Cazzato et al.~\cite{cazzato2020look}. To date, however, no comprehensive review has examined the recent trends in learning from less supervision. Moreover, all of the existing reviews focus only on gaze estimation and ignore significant works in eye segmentation, gaze zone estimation, gaze trajectory prediction, gaze redirection, and unconstrained gaze estimation in single and multiperson setting. The contributions of the paper are summarized below:
\begin{enumerate}[topsep=1pt,itemsep=0pt,partopsep=1ex,parsep=1ex,leftmargin=*]

\item \textbf{A comprehensive review of automated gaze analysis.} 
We categorize and summarize existing methods by considering data capturing sensors, platforms, popular gaze estimation tasks in computer vision, level of supervision and learning paradigm. The proposed taxonomies aim to help researchers to get a deeper understanding of the key components in gaze analysis.

\item \textbf{Different popular tasks under one framework.}
To the best of our knowledge, we are the first to put different eye and gaze related popular tasks under one framework. Apart from gaze estimation, we consider gaze trajectory, gaze zone estimation and gaze redirection tasks.


\item \textbf{Applications.} We explore major applications of gaze analysis using computer vision i.e. Augmented and Virtual Reality~\cite{patney2016perceptually,clay2019eye}, Driver Engagement~\cite{ghosh2020speak2label,vora2018driver} and Healthcare~\cite{harezlak2018application,kempinski2016system}. 

\item \textbf{Privacy Concerns.} We also provide a brief review of the privacy concerns of the gaze data and its possible implications.

\item \textbf{Overview of open questions and potential research directions.} We review several issues associated with the current gaze analysis frameworks (i.e. model design, dataset collection, etc.) and discuss possible future research directions.
\end{enumerate}

\begin{figure*}[t]
    \centering
    \includegraphics[width=\linewidth]{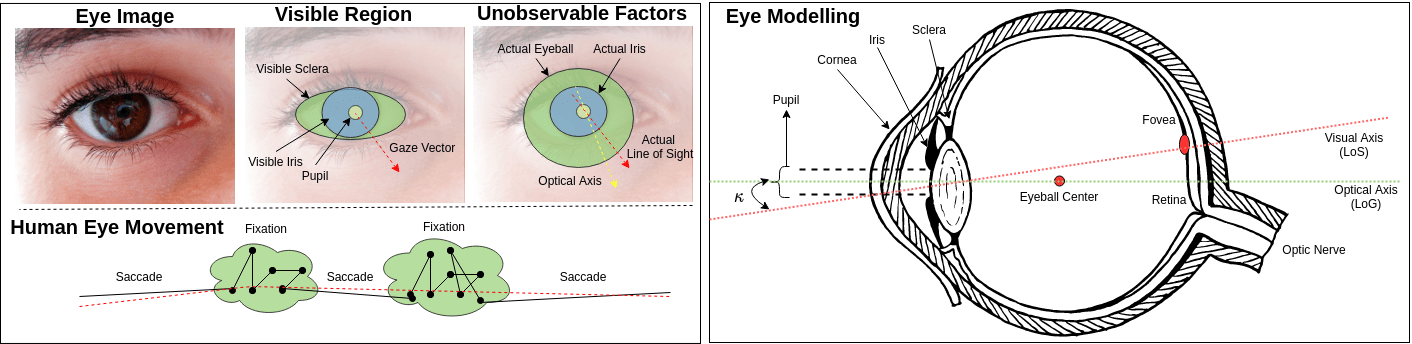}
    \caption{\textbf{Top Left:} Overview of the human visual system, eye modelling and eye movement. For computer vision based automated gaze analysis, we consider an image containing eyes (left) as input. Thus, such methods analyze the visible eye regions (middle) and predict the 2-D/3-D gaze vector as output. However, there are unobservable factors by which we can predict the true gaze direction (right) which requires person-specific information and other factors~\cite{park2020representation}. \textbf{Bottom Left:} Apart from static image based gaze estimation, the dynamic eye movement is another line of research in computer vision that provides cues regarding human behavioural traits. \textbf{Right:} The actual modelling of gaze with respect to eye anatomy. We only highlight the relevant parts i.e. pupil, cornea, iris, sclera, fovea, LOS and LOG. The angle between LOG and LOS is called angle of kappa ($\kappa$).} 
    \label{fig:eye_anatomy}
    \vspace{-5mm}
\end{figure*}



\section{Preliminaries}
\label{sec:priliminary}

The human visual system is a complex cognitive process. As such, understanding and modelling human gaze has become a fundamental research problem in psychology, neurology, cognitive science, and computer vision. To lay the foundations for this review, below, we provide brief descriptions of the \textit{Human Visual System and Eye Modelling} (Sec.~\ref{sec:eye_modelling}), \textit{Eye movements} (Sec.~\ref{sec:eye_movements}), \textit{Problem Settings in Automated Gaze Analysis} (Sec.~\ref{sec:problem_setting}) and the associated \textit{ Challenges} (Sec.~\ref{sec:challenges}).

\subsection{Human Visual System and Eye Modelling} \label{sec:eye_modelling}
Computer vision based human visual perception methods estimate gaze quantitatively from image or video data. These methods analyze the visible region of the eyes (the iris and sclera, see Fig.~\ref{fig:eye_anatomy}), and attempt to approximate the unobservable features of the eyes (which are integral to determining gaze direction). These approximations can be based on models of the human eyes, derived from movement patterns over time, or learned via representation learning over large scale data. For gaze estimation, we typically approximate the eye as a sphere with a radius of 12-13mm. Subsequently, we model gaze direction with respect to the optical axis, also called the \textit{line of gaze (LoG)}, or the visual axis, the \textit{line of sight (LoS)} (see Fig.~\ref{fig:eye_anatomy}right). 
The line of gaze (LoG) connects the pupil, cornea and eyeball center. 
Conversely, the line of sight (LoS) is the line connecting the fovea and center of the cornea. Generally, the LoS is considered as the \textit{true direction of gaze}. The intersection point of the visual and optical axis is called the \textit{nodal point of eye} (anatomically the cornea center), which typically encodes a subject dependent angular offset. This individual offset is the main motivation behind having subject dependent calibration for gaze tracking devices. According to prior studies~\cite{carpenter1988movements,guestrin2006general}, the fovea is located around 4-5\textdegree\ horizontally and 1.5\textdegree\  vertically below the optical axis. Across a broader population, this can vary up to 3\textdegree\ between subjects~\cite{guestrin2006general}. Additionally, head-pose also plays an important role in gaze analysis. The coarse gaze direction of a subject can be determined by the position (in 3-D coordinate) and orientation (Euler angles) of the headpose~\cite{hansen2009eye}. Most of the time, the combined direction of LoS and head pose provide information about where the person is looking.  

\subsection{Eye Movements and Foveal Vision} \label{sec:eye_movements}
We perceive our environment through eye movements. These movements can be voluntary or involuntary, and help us to acquire, fixate, and track visual stimuli (see Fig.~\ref{fig:eye_anatomy}). Eye movements are divided into three primary categories: saccades, smooth pursuits, and fixations.


\noindent \textbf{Saccades.} Saccades are rapid and reflexive eye movements, primarily used for adjustment to a new location in the visual environment. It can be executed voluntarily or involuntarily, as a part of optokinetic measure~\cite{duchowski2017eye}. Saccades typically last for between 10 and 100 ms.  

\noindent \textbf{Smooth Pursuits.} Smooth pursuit occur while tracking a moving target. This involuntary action depends on the range of the target's motion as astonishingly, human eyes can follow the velocity of a moving target to some extend. 

\noindent \textbf{Fixations (Microsaccades, Drifts, and Tremors).} Fixations are eye movements in which the focus of attention stabilizes over a stationary object of interest. Fixations are characterized by three types of miniature eye movements: \textit{tremor, drift} and \textit{microsaccades}~\cite{duchowski2017eye}. During Fixations, the miniature eye movements occur due to the noise present in the control system to hold gaze steadily. This noise occurs in the area of fixation, around 5\textdegree\ visual angle. For simplification of the underlying natural process, this noise is ignored during fixation. 

\noindent \textbf{Foveal Vision.} The \textit{fovea centralis} region of human eye is responsible for the perception of sharp and high resolution human vision. In order to perceive the environment, it is necessary to direct the foveal vision to select region of interest (the process is termed as `foveation'). This sharp foveal vision decays rapidly within the range of 1-5\textdegree. Beyond this limit, human vision is blurred, and low resolution. This is termed as \textit{peripheral vision}. Our peripheral vision plays an important role in our overall visual experience, especially for motion detection. On an abstract level, our visual perception is the result of our brains merging our foveal and peripheral vision. 

\subsection{Gaze Estimation: Problem Setting} \label{sec:problem_setting}
\begin{figure*}[t]
    \centering
    \includegraphics[width=\linewidth]{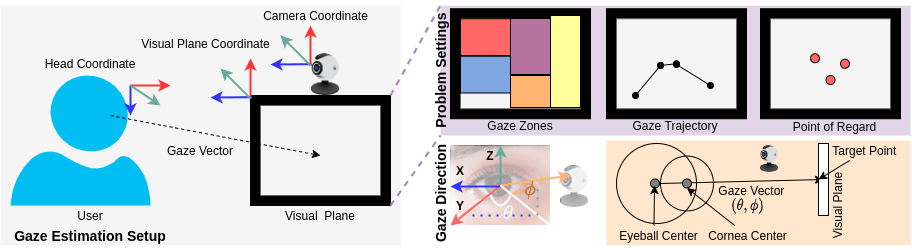}
    \caption{Overview of gaze estimation setups (See Sec.~\ref{sec:problem_setting} for more details). A traditional gaze analysis setup considers the effect of head, visual plane and camera coordinates. The gaze analysis tasks include gaze zone, point of regard, gaze trajectory estimation etc. (See Sec.~\ref{sec:task}) The gaze vector is defined by the angles ($\theta, \phi$) in polar co-ordinate systems as shown in the gaze direction part.} 
    \label{fig:problem_setup}
\end{figure*}

The main task of gaze estimation is to determine the line of sight of the pupil. Fig.~\ref{fig:problem_setup} depicts a typical visual sensor based real-time gaze estimation setup consisting of user, data capturing sensor(s) and visual plane. The main calibration factors in this setting are:  
\begin{itemize}[topsep=1pt,itemsep=0pt,partopsep=1ex,parsep=1ex,leftmargin=*]
    \item Estimation of \textit{camera calibration} parameters, which include both intrinsic and extrinsic camera parameters.
    \item Estimation of \textit{geometric calibration} parameters, which include the relative position of the camera, light source and screen.
    \item Estimation of \textit{personal calibration}, which include headpose and eye-specific parameters such as cornea curvature, the nodal point of the eye, etc.
\end{itemize}
In some of the applications, the calibration parameters are estimated in task-specific settings. For example, users are requested to fixate their gaze to some pre-defined points for calibration. Similarly, user-specific information is registered once in the devices for subject-specific calibrations. With the advances in computer vision and deep learning, nowadays gaze estimation techniques are developed with appearance based features and do not require an explicit calibration step. For example, \textit{CalibMe}~\cite{santini2017calibme} is a fast and unsupervised calibration technique for gaze trackers, designed to overcome the burden of repeated calibration.
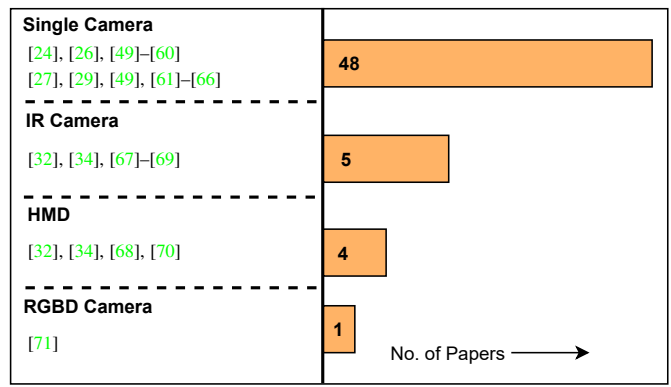
\begin{figure}[t]
  \centering
      \includegraphics[width=0.99 \linewidth]{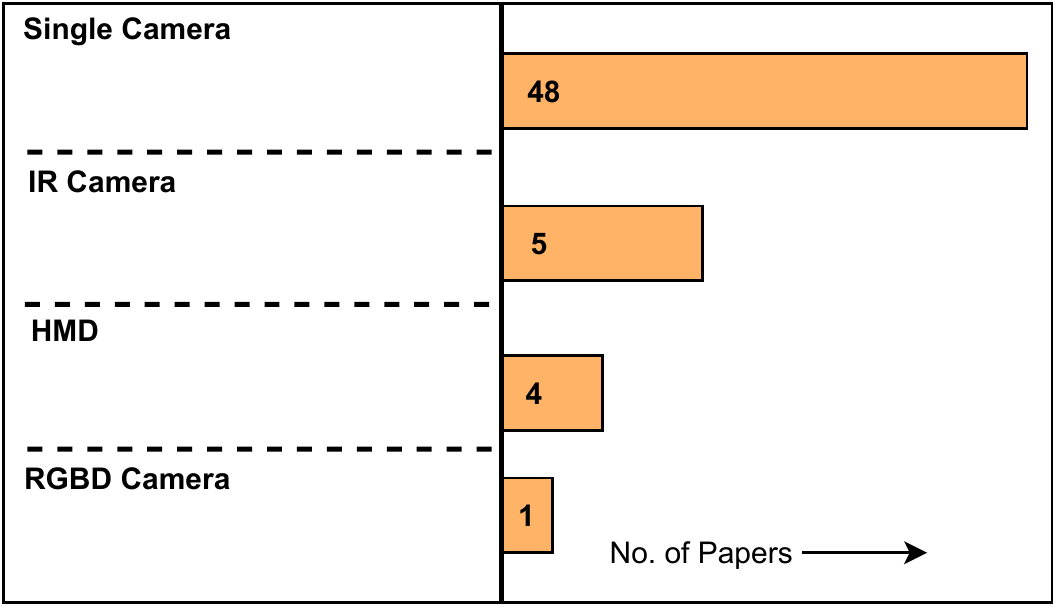}
      \put(-245,125){\scriptsize{\cite{zhang2019evaluation,zhang2015appearance,zhang2017mpiigaze,zhang2017s,zhu2017monocular,cui2017specialized,park2018deep,FischerECCV2018,cheng2018appearance,palmero2018recurrent,jyoti2018automatic,liu2019differential,cheng2020coarse,zhang2020learning}}}
      \put(-245,115){\scriptsize{\cite{yu2018deep,chong2018connecting,zhang2018training,zhang2019evaluation,park2019few,chen2018appearance,zhou2019learning,wang2019neuro,dubey2019unsupervised}}}
      \put(-245,85){\scriptsize{\cite{wu2019eyenet,kim2019nvgaze,garbin2019openeds,palmero2020openeds2020,xia2007ir}}}
      \put(-245,50){\scriptsize{\cite{tonsen2017invisibleeye,kim2019nvgaze,garbin2019openeds,palmero2020openeds2020}}}
      \put(-245,15){\scriptsize{\cite{lian2019rgbd}}}
\caption{The plot shows the popularity of different data capturing devices across research articles over the past 10 years. Here, HMD: Head Mounted Device, RGBD: RGB Depth camera. 
}
\label{fig:data_sensors}
\vspace{-5mm}
\end{figure}

\noindent \textbf{Role of Data Capturing Sensors.} 
Visual stimuli provide valuable information for computer vision based gaze estimation techniques. A trade-off of widely used sensors is mentioned in Fig.~\ref{fig:data_sensors}. These data capturing sensors are mainly divided into two categories: \textit{Intrusive} and \textit{Non-intrusive}. An array of methods used specialized hardware which requires physical contact with human skin or eyes are termed as  \textit{intrusive sensors}. The widely used intrusive sensors are head-mounted devices (HMD), electrodes, or scleral coils~\cite{xia2007ir,tsukada2011illumination}. These devices may cause unpleasant user experience, and the accuracy of these systems depends on the tolerance, accurate subject-specific calibration and other factors of the devices. On the other hand, data capturing devices that do not require physical contact~\cite{leo2014unsupervised} are termed \textit{non-intrusive} sensors. Mainly, RGB, RGBD and IR cameras fall under this category. These methods face several challenges, which include partial occlusion of the iris by the eyelid, varying illumination condition, head pose, specular reflection in case the user wears glasses, the inability to use standard shape fitting for iris boundary detection, and other effects including motion blur and over saturation of images~\cite{leo2014unsupervised}. To deal with these challenges, most of the existing gaze estimation methods have been performed under constrained environments like constrained head pose, controlled illumination conditions, and camera angle. Among all of the aforementioned factors, pupil visibility plays an important role as robust gaze estimation needs accurate pupil-center localization. Fast and accurate pupil-center localization is still a challenging task~\cite{gou2017joint}, particularly for images with low resolution. A trade-off of widely used sensors are mentioned in Fig.~\ref{fig:data_sensors}.

\noindent \textbf{Role of Headpose.} Gaze estimation is a challenging task due to eye-head interplay. Head-pose plays the most important role in gaze estimation. The gaze direction of a subject is determined by the combined effect of position and orientation of head pose and eyeball. One can change gaze direction via eyeball and pupil movement by maintaining stationary or dynamic head-pose or by moving both. Usually, this process is subject dependent. People adjust their head-pose and gaze to maintain a comfortable posture. Thus, the gaze estimation task needs to consider both gaze and head-pose at the same time for inference. As a result of this, it is more common to consider head-pose information in the gaze estimation methods implicitly or explicitly~\cite{zhang2015appearance,zhang2017mpiigaze,pi2019task}.

\noindent \textbf{Role of Visual Plane.}
The visual plane is the plane containing the gaze target point i.e. where the subject is looking, which is often termed as Point of Gaze (PoG). 
The distance between the user and the visual plane varies a lot in a real-world setting. Thus, recent deep learning based methods do not rely on the distance or placement of the visual plane. The most common gaze analysis setup uses a RGB camera placed at 20 - 70 cm from the user in unconstrained setting i.e. without any invasive sensors or fixed setup. In different real-world settings, the visual plane could be desktop ($\sim$ 60cm), mobile phone ($\sim$ 20cm), car ($\sim$ 50cm) etc. An overview is presented in Table \ref{tab:platforms}.

\begin{table}[t]
\caption{Attributes of different platforms widely used in gaze analysis. Here, Dist.: distance (in cm), VA: viewing Angle (in \textdegree), HMD: Head Mounted Devices, FV: Free Viewing, UC: User Condition, ET: External Target.} 
\label{tab:platforms} 
\centering
\begin{tabular}{l|c|c|l|l}
\toprule[0.4mm]
\rowcolor{mygray}
\multicolumn{1}{l|}{\textbf{Platform}} & \multicolumn{1}{c|}{\textbf{Dist.}} & \multicolumn{1}{c|}{\textbf{VA}} & \multicolumn{1}{l|}{\textbf{UC}} & \multicolumn{1}{c}{\textbf{Papers}}  \\ \hline \hline

\begin{tabular}[c]{@{}l@{}}Desktop,\\TV Panels\end{tabular}  & \begin{tabular}[c]{@{}l@{}}30-50, \\200-500 \end{tabular}                                                             & \begin{tabular}[c]{@{}l@{}}$\sim$ 40 \textdegree, \\   40\textdegree-60\textdegree      \end{tabular}                                                     & \begin{tabular}[l]{@{}l@{}} Static,\\ Sitting,\\ Upright \end{tabular}          &  \begin{tabular}[l]{@{}l@{}}\cite{zhang2019evaluation,zhang2015appearance,zhang2017mpiigaze,zhang2017s,zhu2017monocular,cui2017specialized,park2018deep,FischerECCV2018,cheng2018appearance,palmero2018recurrent,jyoti2018automatic} \\~\cite{yu2018deep,chong2018connecting,zhang2018training,zhang2019evaluation,park2019few,chen2018appearance,zhou2019learning,wang2019neuro}\\~\cite{lian2018multiview,lian2019rgbd,liu2019differential,cheng2020coarse,zhang2020learning}  \end{tabular}                                                \\ \hline
HMD & 2-5                                                            & 55\textdegree-75\textdegree                                                               & \begin{tabular}[l]{@{}l@{}} Independent\\(Leanback,\\ Sitting,\\ Upright) \end{tabular}                                                            &  \begin{tabular}[l]{@{}l@{}}~\cite{jha2018probabilistic,hu2020dgaze,garbin2019openeds,palmero2020openeds2020} \end{tabular}  \\ \hline
Automotive & 50                                                            & 40\textdegree-60\textdegree                                                               & \begin{tabular}[l]{@{}l@{}} Mobile,\\ Sitting,\\ Upright \end{tabular}                                                           &   \begin{tabular}[l]{@{}l@{}}~\cite{ghosh2020speak2label,tawari2014driver,tawari2014robust,vasli2016driver,fridman2015driver,fridman2016owl,choi2016real,lee2011real} \end{tabular} \\ \hline
Handheld & 20-40                                                           & 5\textdegree-12\textdegree                                                               & \begin{tabular}[l]{@{}l@{}} Leanfwd,\\ Sitting,\\ Standing,\\Mobile \end{tabular}                                                             &  \begin{tabular}[l]{@{}l@{}}~\cite{krafka2016eye,zhang2018training,he2019device,huang2015tabletgaze,guo2019generalized,bao2021adaptive} \end{tabular} \\ \hline
ET/ FV & --                                                           & --                                                              & \begin{tabular}[l]{@{}l@{}} Leanfwd,\\ Sitting,\\ Standing,\\ Upright \end{tabular}                                                             &  \begin{tabular}[l]{@{}l@{}}~\cite{gaze360_2019,dubey2019unsupervised} \end{tabular} \\
\bottomrule[0.4mm]
\end{tabular}
\vspace{-5mm}
\end{table}

\subsection{Challenges}
\label{sec:challenges}

\noindent \textbf{Data Annotation.} 
Generally, deep learning based methods require large amount of annotated data for generalized representation learning. Curating large scale annotated gaze datasets is non-trivial~\cite{gaze360_2019,zhang2015appearance,huang2015tabletgaze}, time consuming and requires expensive equipment setup. Current dataset recording paradigms via wearable sensors may lead to uncomfortable user experience and it require expert knowledge. Another common aspect of the current datasets is the constrained environment in which they are recorded  (For example, CAVE~\cite{smith2013gaze} dataset is recorded on indoor environment with headpose restricted). Recently, a few datasets \cite{gaze360_2019,huang2015tabletgaze} have been proposed to address this gap by recording in unconstrained indoor and outdoor environments. Another challenge associated with data annotation is participant's cooperation. However, it is assumed that participants fixate their gaze as per the given instructions~\cite{gaze360_2019,zhang2015appearance,huang2015tabletgaze}. Despite these attempts~\cite{gaze360_2019,zhang2015appearance,huang2015tabletgaze}, data annotation still remains complex, noisy and time-consuming. Self/weakly/un-supervised learning paradigms~\cite{dubey2019unsupervised,yu2019unsupervised} could be helpful to address the dataset creation and annotation challenges.
    
\noindent \textbf{Subjective Bias.}
Another challenge for the automatic gaze analysis method is subjective bias. Individual differences in the nodal points of human eyes makes automatic and generic gaze analysis way more difficult.
In an ideal scenario, any gaze analysis method should encode rich features corresponding to eye region appearance, which provides relevant information for gaze analysis. To address this challenge, few-shot learning based approach has been widely adapted ~\cite{park2019few,yu2019improving}, where the motivation is to adapt to new subject with minimum subject specific information. Another way to deal with the subjective bias is combining classical eye model based approaches with geometrical constraints~\cite{wang2018hierarchical} as this approach has the potential to generalize well across subjects. 

    
    
\noindent \textbf{Eye Blink.}
Blinks are an involuntary and periodic motion of the eyelids. They pose a challenge for gaze analysis, as blinks result in missed frames of data. A few recent works~\cite{gaze360_2019,ghosh2020speak2label} assume that the head pose information is a suitable replacement for gaze during blinking based on a common line of sight between a subject's headpose and gaze. However, it is noted that a large shift in the gaze is possible after subject re-opens their eyes. To simplify the situation, some gaze analysis methods ignore eye-blink data (e.g.,~\cite{huang2015tabletgaze,jyoti2018automatic}) and some treat blinks as a separate class of data (.e.g,~\cite{vora2017generalizing,vora2018driver}). 
A possibility for real world deployment of such a system is generating gaze labels by interpolating from neighbouring frames' labels when blinks are detected~\cite{ghosh2020speak2label}.
    
\noindent \textbf{Data Attributes.}
Several factors, such as eye-head interplay, occlusion, blurred image, and illumination can influence the performance of a gaze analysis model. The presence of any subset of these attributes can degrade the performance of a system~\cite{gaze360_2019,huang2015tabletgaze}. Many methods use face alignment~\cite{zhang2015appearance,krafka2016eye} and 3-D head pose estimation~\cite{zhang2015appearance} as a pre-processing step. However, face alignment on images captured in an unconstrained environment based images may introduce noise in a system. To overcome this, recent approaches~\cite{gaze360_2019,zhang2020eth,jyoti2018automatic,dubey2019unsupervised} avoid these pre-processing steps and show increase in gaze prediction performance.

Another critical challenge in gaze estimation is eye-head interplay. Prior studies generally address this issue via implicit training~\cite{krafka2016eye,zhang2017everyday} or provide the head pose information separately as a feature~\cite{zhang2015appearance}. Similarly, it is challenging to estimate gaze under partial occlusion. When the head's yaw rotation is greater than 90\textdegree, one side of the face becomes occluded w.r.t. the camera. A few prior works~\cite{krafka2016eye,zhang2015appearance} avoid these scenarios by disregarding these frames. Kellnhofer et al.~\cite{gaze360_2019}, however, argue that when the head yaw angle is in the range 90\textdegree - 135\textdegree, the partial visibility still provides relevant information about the gaze direction. This study also proposes quantile regression via pinball loss to mitigate the effect of partial occlusion in training data in terms of uncertainty. Despite all of these attempts, gaze estimation still remains challenging in presence of these attributes. There is still have scope to eliminate the effects of these attributes and make the gaze analysis model more robust for the real-world deployment.

\noindent \textbf{Application Specific Challenges.} 
Gaze analysis also has application-specific requirements, for example, coarse or fine gaze estimation in AR, VR, Robotics, egocentric vision and HCI. Thus, a working algorithm behind any eye-tracking devices needs to fit in the application environment. 

\section{Gaze Analysis in Computer Vision} \label{sec:task}
We provide a breakdown of different gaze analysis tasks for vision based applications. Any statistical gaze modeling mainly estimates the relation between the input visual data and the point of regard/gaze direction. 


\noindent \textbf{2-D/3-D Gaze Estimation.}
Most of the existing studies consider gaze estimation as either the gaze direction in 3-D space or as the point of regard in 2-D/3-D coordinates (see Fig.~\ref{fig:problem_setup}). We can divide the gaze estimation methods into the following types:

\noindent \textit{1) Geometric Methods:} These geometric methods compute a gaze direction from the geometric model of the eye (see Fig.~\ref{fig:eye_anatomy}), where the anatomical structure of the eye is considered to get the 3-D gaze direction or gaze vector. These methods were widely used in prior to more deep learning approaches~\cite{hansen2009eye}.
These recent deep learning based approaches implicitly model these geometric parameters during the learning process, and, as such, do not explicitly require the often noisy subject specific parameters, such as cornea radii, cornea center, angles of kappa (i.e. Refer $\kappa$ in Fig.~\ref{fig:eye_anatomy}), iris radius, the distance between the pupil center and cornea center, etc. 

\noindent \textit{2) Regression Methods:} Regression based methods~\cite{morimoto2005eye,zhang2015appearance,park2018deep,yu2019unsupervised} map visual stimuli (image or image-related features) to gaze coordinates or gaze angles in 2-D/3-D. The output mapping is application-specific. For example, such techniques are often used to map 2-D/3-D gaze coordinate mainly maps people's focus of attention to the screen coordinates (for human-computer interaction based applications such as engagement or attention monitoring). Regression based methods can be divided into two types: the \textit{parametric} and \textit{non-parametric} approaches. Parametric approaches (e.g.,~\cite{morimoto2005eye,yu2019unsupervised}) assume gaze trajectories as a polynomial, where the task is to estimate the parameters of the polynomial equation. \textit{Non-parametric} approaches directly work on the mappings in spite of calculating the intersection between the gaze direction and gazed object explicitly~\cite{zhang2015appearance,park2018deep,jyoti2018automatic}. The recent deep learning based approaches are  non-parametric~\cite{park2018deep,park2018learning,yu2019unsupervised,FischerECCV2018,gaze360_2019}.

\noindent \textbf{Trajectory Prediction.} Gaze estimation has potential applications in AR/VR especially in Foveated Rending (FR) and Attention Tunneling (AT), where the future eye trajectory prediction is highly desirable. To meet this requirement, a new research direction (i.e. future gaze trajectory prediction) has been recently introduced~\cite{palmero2020openeds2020}. Here, possible future gaze locations can be estimated based on the prior gaze points, content of the visual plane or their combination. Thus, the problem statement can be formulated as follows: given $n$ number of prior gaze points, the algorithm will predict the $m$ future frames' gaze direction in a user-specific setting.


\noindent \textbf{Gaze Zone.} In many gaze estimation based applications such as driver gaze~\cite{ghosh2020speak2label,vora2018driver,vora2017generalizing,dubey2019unsupervised}, gaming platforms~\cite{corcoran2012real}, website designing~\cite{chu2009using}, etc., the exact position or angle of the line of sight of the pupil is not required. Thus, a gaze zone approach is utilized in these cases for estimation. Here, the gaze zone refers to an area in 2-D or 3-D space. For example, in a simplistic driver gaze zone estimation, the driver could be looking straight ahead, at the steering wheel, at the radio, or at the mirrors.
Similarly, another example is detecting more visually salient zone/region during website designing~\cite{chu2009using}.

\noindent \textbf{Gaze Redirection.}
Due to the challenges in different posed gaze conditions, generation on the go is gaining popularity~\cite{chen2020mggr,palmero2020openeds2020}. It aims to capture subject-specific signals from a few eye images of an individual and generate realistic eye images for the same individual under different eye states (gaze direction, camera position, eye openness etc.). The gaze redirection can be performed in both controlled and uncontrolled way~\cite{zheng2020self,chen2020mggr,chen2021coarse}. Apart from these, eye rendering is another research direction to generate realistic eye given the appearance, gaze direction of a person. It has potential applications in virtual agents, social robotics, behaviour generation and in the animation industry~\cite{ruhland2015review}.

\noindent \textbf{Unconstrained Gaze Estimation.}
Gaze estimation in an unconstrained setting can be divided into two types:

\noindent \textit{1) Single Person Setting:} In webcam or RGB camera based gaze estimation approaches, geometric model based eye tracking~\cite{baltruvsaitis2016openface,baltrusaitis2018openface} is typically used, as it is fast and does not require training data. 
At the same time, however, it relies on accurate eye location and key points detection, which is hard to achieve in real-world environments. Deep learning based methods~\cite{king2009dlib,park2018learning} have eliminated this issue to some extent, however, it still remains a challenge as it does not generalize well in different settings. 

\noindent \textit{2) Multi-Person Setting:} In unconstrained multi-person settings, it is very difficult to track the eyes. For example, in a social interaction scenario, understanding the gaze behaviour of each person provides important cues to interpret social dynamics~\cite{mueller18_etra}. To this end, a new research direction is introduced where the problem is defined as whether the people are Looking At Each Other (LAEO) in a given video sequence~\cite{marin2019laeo,marin2021pami,Kothari_2021_CVPR}. Similarly, gaze communication~\cite{fan2019understanding} and GazeOnce~\cite{Zhang_2022_CVPR} are another line of research aligned with this field.

\noindent \textbf{Visual Attention Estimation.}
Human visual attention estimation is another line of research which mainly focuses on \textit{where the person is looking} irrespective of eyes visibility. The popular sub-tasks in this direction are gaze following~\cite{NIPS2015_ec895663,chong2020detecting,chong2018connecting,Tu_2022_CVPR,Wang_2022_CVPR_GaTector}, gaze communication~\cite{fan2019understanding}, human attention in goal-driven environments~\cite{yang2020predicting} and categorical visual search~\cite{zelinsky2019benchmarking}, visual scan-path analysis in visual question answering~\cite{chen2021predicting}, and naturalistic environment~\cite{kummerer2021state,Bao_2022_CVPR}. These methods are mostly driven by saliency in the scene, head orientation, or any other task at hand. Visual attention based approaches have the potential to localize the gaze target directly from scene information which in turn enhances the scalability of naturalistic gaze behaviour patterns.




\section{Gaze Analysis Framework} \label{sec:framework}
\begin{figure}[t]
    \centering
    \includegraphics[width=\linewidth]{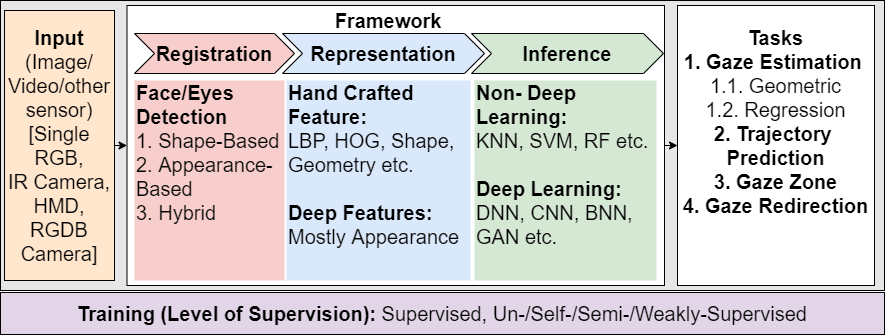}
    \caption{A generic gaze analysis framework has different components including registration, gaze representations and inference. Although in the deep learning based approaches, there is a high overlap between the representation and inference module. Refer Sec.~\ref{sec:framework} for more details.}
    \label{fig:overview}
    \vspace{-5mm}
\end{figure}

We break down a gaze analysis framework into its fundamental components (Fig.~\ref{fig:overview}) and discuss their role in terms of \textit{eye detection and segmentation} (Sec.~\ref{sec:eye_detection}), \textit{Network Architecture} (Sec.~\ref{sec:pipelines}) and \textit{Level of Supervision} (Sec.~\ref{sec:supervision}).

\subsection{Eye Detection and Segmentation}\label{sec:eye_detection}
Eye registration is the first stage of gaze analysis and requires detection of the eye and the relevant regions of interest.

\noindent \textbf{Eye Detection Methods.}
The main aim of the eye detection algorithms is to accurately identify the eye region from an input image. Eye detection algorithms need to operate in challenging conditions such as occlusion, eye openness, variability in eye size, head pose, illumination, and viewing angle, while balancing the trade-off in appearance, dynamic variation and computational complexity. Prior works on eye detection can be divided into three categories: shape based~\cite{hansen2005eye}, appearance based~\cite{liang2013appearance,wang2016hybrid,zhang2019evaluation,baltrusaitis2018openface} and hybrid method~\cite{wang2016hybrid}. The most popular libraries for eye and facial point detection are \href{https://github.com/davisking/dlib}{Dlib}~\cite{king2009dlib} \href{https://github.com/cmusatyalab/openface}{OpenFace}~\cite{baltrusaitis2018openface,baltruvsaitis2016openface}, \href{https://github.com/kpzhang93/MTCNN\_face\_detection\_alignment}{MTCNN}~\cite{zhang2016joint}, \href{https://github.com/Tencent/FaceDetection-DSFD}{Duel Shot Face Detector}~\cite{li2018dsfd}, \href{https://github.com/JDAI-CV/FaceX-Zoo}{FaceX-Zoo}~\cite{wang2021facex}. 

The pupil and iris region of the eye is usually darker than the sclera which provides an important cue to differentiate or localize the pupil. The pupil center localization use dedicated and costly devices~\cite{xia2007ir,tsukada2011illumination}, which requires person-specific pre-calibration. To overcome this limitation, the deep learning based pupil localization methods use ensembles of randomized trees~\cite{markuvs2014eye}, local self similarity matching~\cite{leo2014unsupervised}, adaptive gradient boosting~\cite{tian2016accurate}, hough regression forests~\cite{kacete2016real}, deep learning based landmark localization models~\cite{park2018learning,king2009dlib}, heterogeneous CNN models~\cite{choi2019accurate}, etc. In prior literature, the choice of eye registration process is influenced by the correlation between the input image and the learning objective of the proposed method. Apart from this, the trade-off between the accuracy of eye localization and running time complexity of the algorithm is optimized in a task specific way. In this context, \textit{OpenFace} and \textit{Dlib} are the most popular. Moreover, the choice of eye/face registration process may also depend on their ability to detect eye components in different challenging real world conditions. 

\noindent \textbf{Eye Segmentation.}
The main task of eye segmentation is pixel-wise or region-wise differentiation of the visible eye parts. In general, the eye region is divided into three parts: sclera (the white region of the eyes), iris (the colour ring of tissue around the pupil) and pupil (the dark iris region). Prior studies~\cite{sankowski2010reliable,radu2015robust,das2017sserbc,lucio2018fully} on eye segmentation mainly explore to segment the iris and sclera region. Few studies~\cite{garbin2019openeds,palmero2020openeds2020} include the pupil region in the segmentation task as well. Eye segmentation is widely used in the biometric systems~\cite{das2013sclera} and prior for synthetic eye generation~\cite{chaudhary2019ritnet}. 

\noindent \textbf{Eye Blink Detection.}
Eye blinks are the involuntary and periodic activity that can help to judge the cognitive activity of a person (e.g. driver's fatigue~\cite{pandey2021real}, lie detection~\cite{monaro2020using}). KLT trackers and various sensors are also widely used to get the eye motion information to track eye blink~\cite{drutarovsky2014eye}. The existing eye blink detection approaches aim to solve a binary classification problem (blink/no blink) either in a \textit{heuristic based} or \textit{data-driven} way. The \textit{heuristic based} approaches mainly include motion localization~\cite{drutarovsky2014eye} and template matching~\cite{krolak2012eye}. As these methods are highly reliable on pre-defined thresholds, they could be sensitive to subjective bias, illumination and head pose. To overcome this limitation, the \textit{data-driven} approaches infer on the basis of appearance based temporal motion features~\cite{cech2016real,drutarovsky2014eye} or spatial features~\cite{daza2020mebal}. In \textit{hybrid approach}~\cite{hu2019towards}, multi-scale LSTM based framework is used to detect eye blink using both spatial and temporal information.

\begin{figure*}[t]
  \centering
      \includegraphics[width=0.99 \linewidth]{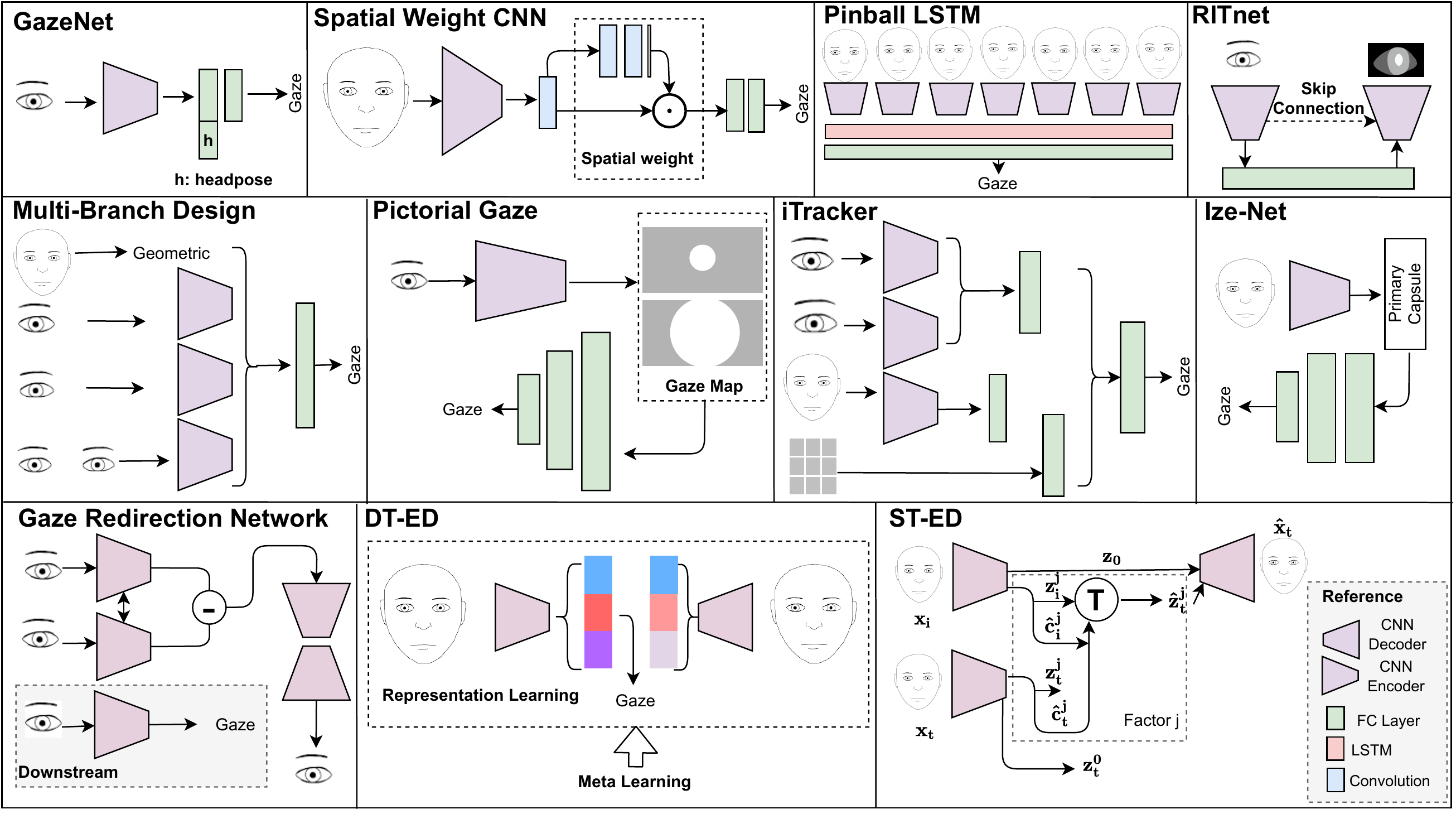}
      \put(-473,276){\scriptsize{\cite{zhang2015appearance}}}
      \put(-340,270){\scriptsize{\cite{zhang2017s}}}
      \put(-168,279){\scriptsize{\cite{gaze360_2019}}}
      \put(-65,279){\scriptsize{\cite{chaudhary2019ritnet}}}
      \put(-424,210){\scriptsize{\cite{jyoti2018automatic}}}
      \put(-322,210){\scriptsize{\cite{park2018deep}}}
      \put(-205,210){\scriptsize{\cite{krafka2016eye}}}
      \put(-60,210){\scriptsize{\cite{dubey2019unsupervised}}}
      \put(-405,95){\scriptsize{\cite{yu2019unsupervised}}}
      \put(-359,102){\scriptsize{\cite{park2019few}}}
      \put(-175,102){\scriptsize{\cite{zheng2020self}}}
\caption{A brief overview of different pipelines used for gaze analysis tasks. Refer Sec.~\ref{sec:pipelines} for more details of the networks.}
\label{fig:pipelines}
\vspace{-5mm}
\end{figure*}
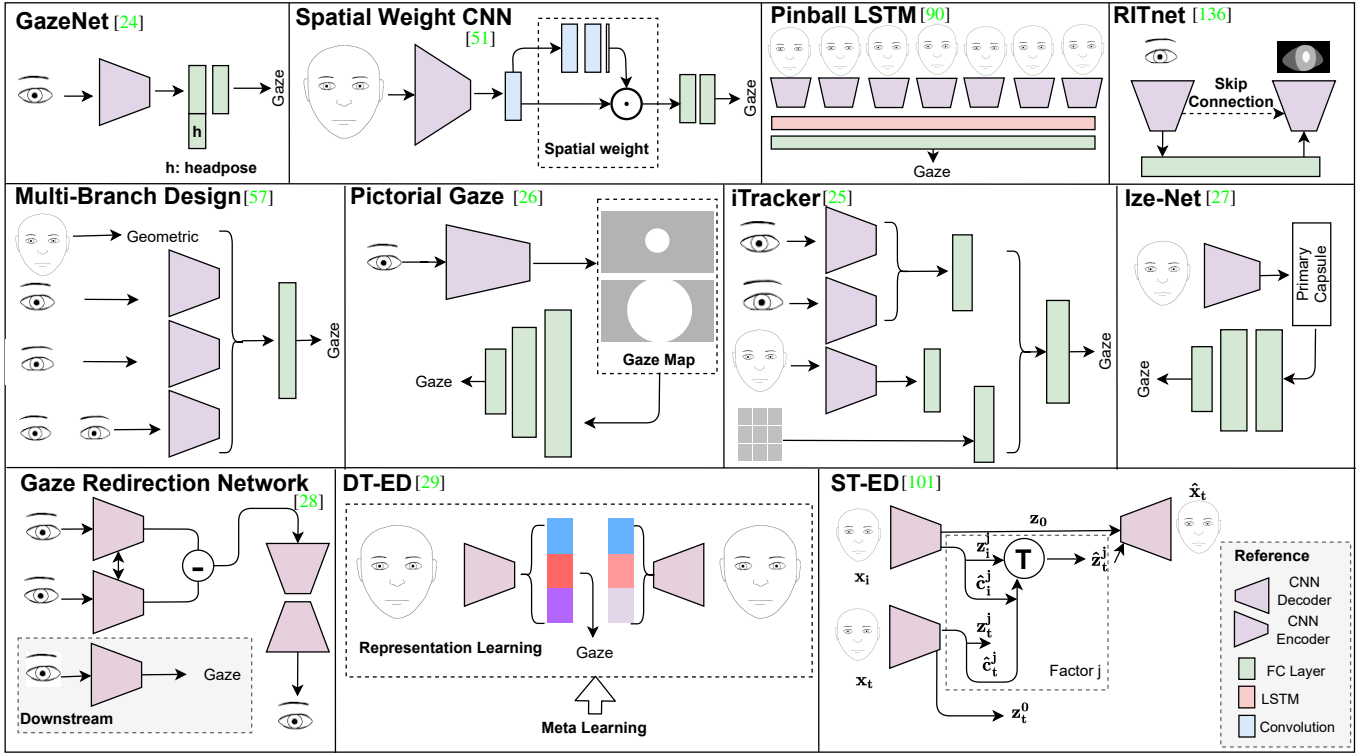

\subsection{Representative Deep Network Architectures} \label{sec:pipelines}
In this section, we provide a generic formulation and representation of gaze analysis. Given an RGB image $\mathbf{I}\in \mathbb{R}^{W\times H \times3} $, a deep learning based model mapped it to task-specific label space. The input RGB image is usually the face or eye regions. Based on the primary network architectures adopted in the literature, we classify the models into the following categories: \textit{CNN based, Multi-Branch network based, Temporal based, Transformer Based} and \textit{VAE/GAN based}. An overview is shown in Fig.~\ref{fig:pipelines}. 

\subsubsection{CNN based} Most of the recent solutions adopt a CNN based architecture~\cite{zhang2015appearance,zhang2017s,zhang2017mpiigaze,krafka2016eye,park2018deep,wang2018hierarchical}, which aims to learn end-to-end spatial representation followed by gaze prediction. The adopted model is often a modified version of the popular CNNs in vision (e.g. AlexNet~\cite{dubey2019unsupervised}, VGG~\cite{FischerECCV2018}, ResNet-18~\cite{gaze360_2019,wu2019eyenet}, ResNet-50\cite{zhang2020eth}, Capsule network~\cite{dubey2019unsupervised}). These CNNs learn from a single-stream of RGB images (e.g. face, left or right eye patch)~\cite{zhang2015appearance,zhang2017s}, or multiple streams of information (e.g. face, and eye patches)~\cite{jyoti2018automatic,krafka2016eye}, and prior knowledge based on eye anatomy or geometrical constraints~\cite{park2018deep}. 

\noindent \textbf{GazeNet.} It is the extended version of the first deep learning based gaze estimation method~\cite{zhang2015appearance} which aims to capture low level and high level appearance feature by using convolution operation. GazeNet takes a grayscale eye patch image $\mathbf{I}\in \mathbb{R}^{W\times H} $ as input and maps it to angular gaze vector $\mathbf{g}\in \mathbb{R}^{2} $. As headpose provides relevant features for gaze direction, the headpose vector is also added in the FC layer for better inference (Refer top left image in Fig.~\ref{fig:pipelines}). The Extended version~\cite{zhang2017mpiigaze} is adapted from the VGG network which further boosts the performance. To train these models, the sum of the individual $\ell_2$ losses between the predicted $\mathbf{\hat{g}}$ and actual gaze angle vectors $\mathbf{g}$ is considered. 

\noindent \textbf{Spatial Weight CNN.} It is a full face appearance based gaze estimation method~\cite{zhang2017s} which uses a spatial weighting mechanism for encoding the important locations of the facial image $\mathbf{I}\in \mathbb{R}^{W\times H \times 3} $ via the standard CNN architecture (Refer top row second column image in Fig.~\ref{fig:pipelines}). This weighting mechanism (aka attention) automatically assign more weight to the regions contributing more towards gaze estimation. It includes three additional $1 \times 1$ convolutional layers followed by a ReLU activation. Given a $N \times H \times W$ dimensional activation map ($U$) as input (where $N$, $H$ and $W$ are the number of feature channels, height and width of the output), the spatial weights module learns the weight matrix $W$ from element-wise multiplication of $W$ with the original activation $U$ via the following function: $ W \bigodot U_c,$ across the channel dimensions. Thus, the model learns to assign more weight to the specific regions, which in turn eliminates unwanted noise in the input. For 2-D gaze estimation, the $\ell_1$ distance between the predicted and ground-truth gaze positions in the target screen coordinate system is utilized. Similarly, the $\ell_1$ distance between the predicted and ground-truth gaze angle vectors in the normalized space is used for 3-D gaze estimation.

\noindent \textbf{Dilated Convolution.} Another interesting architecture for gaze estimation is dilated-convolutional layers which preserve spatial resolution while increasing the size of the receptive field without compromising the number of parameters~\cite{chen2018appearance}. It aims to capture the slight change in pixels due to eye movement. Given an input feature map $U$ of kernel size $N \times M \times K$ ($N$: height, $M$:width, $K$:channel with weights $W$ and bias $b$) and dilation rates ($r_1$, $r_2$), the output feature map $v$ can be defined as follows:
$$ v(x, y) = \sum\limits_{k=1}^K \sum\limits_{m=0}^{M-1} \sum\limits_{n=0}^{N-1} u(x + n r_1, y + m r_2, k) w_{nmk} + b $$
The dilated convolution is applied in facial and left/right eye patches before inferring the gaze. For training the network, cross entropy loss is used in the label space. Representation learning via MinENet~\cite{perry2019minenet} also relies on dilated and asymmetric convolutions to provide context to the segmented regions of the eye by increasing the receptive field capacity of the model to learn contextual information.

\noindent \textbf{Bayesian CNN.} Another variant of CNN is Bayesian CNN which is used for robust and generalizable eye-tracking under different conditions~\cite{wang2019generalizing}. Instead of predicting eye gaze using a single trained eye model, it performs eye tracking using an ensemble of models, hence alleviating the over-fitting problem, is more robust under insufficient data, and can generalize better across datasets. Compared to the point based eye landmark estimation methods, the BNN model can generalize better and it is also more robust under challenging real-world conditions. Additionally, the extended version of the BCNN (i.e. the single-stage model to multi-stage, yielding the cascade BCNN) allows feeding the uncertainty information from the current stage to the next stage to progressively improve the gaze estimation accuracy. This could be an interesting area for further study.

\noindent \textbf{Pictorial Gaze.}
Pictorial gaze~\cite{park2018deep} aims to model the relative position of eyeball and iris to get the gaze direction. The network~\cite{park2018deep} consists of two parts: 1) regression from eye patch image to intermediate gazemap followed by 2) regression from gazemap to gaze direction vector $g$ (Refer second row second column image in Fig.~\ref{fig:pipelines}). The gazemap is an intermediate representation of a simple model of the human eyeball and iris in terms of $m \times n$ dimensional image, where, the projected eyeball diameter is $2r = 1.2n$ and the iris centre coordinates ($u_i, v_i$) are as follows: $u_i = \frac{m}{2} - r' sin\ \phi\ cos\ \theta, v_i = \frac{n}{2} - r' sin\ \theta$
where, $r' = r\ cos\ (sin^{-1}\frac{1}{2})$ and gaze direction $g = (\theta, \phi)$. Basically, the iris is an ellipse with major-axis diameter of r and minor-axis diameter of $r |cos\ \theta cos\ \phi|$. The first part is implemented via a stacked hourglass architecture which assumed to encode complex spatial relations including the locations of occluded key points. 
Consequently, for the second part, a DenseNet architecture is used which maps the intermediate gazemap to the gaze vector $\hat{g}$. It is trained via gaze direction regression loss defined as: $||g - \hat{g}||_2$ as well as cross-entropy loss between predicted and ground-truth gazemaps for all pixels.

\noindent \textbf{Ize-Net.} This framework is used for coarse to fine gaze representation learning. Here, the main idea is to learn coarse gaze representation by dividing the gaze locations to gaze zones. Further, the gaze zone is mapped to the finer gaze vector. The proposed network~\cite{dubey2019unsupervised} (Refer second row right image in Fig.~\ref{fig:pipelines}) is a combination of convolution and primary capsule layer. After the convolution layers, the primary capsule layer is appended whose job is to take the features learned by convolution layers and produce combinations of the features to consider face symmetry into account. This network is trained for coarse gaze zone which is fine-tuned for downstream 2-D/3-D gaze estimation.

\noindent \textbf{EyeNet.} consists of modified residual units as the backbone, attention blocks and multi-scale supervision architecture. This network is robust for the low resolution, image blur, glint, illumination, off-angles, off-axis, reflection, glasses and different colour of iris region challenges.

\subsubsection{Multi-Branch network based}
There are several works~\cite{FischerECCV2018,krafka2016eye,jyoti2018automatic} which utilize multiple inputs for better inference. 

\noindent \textbf{iTracker.} The iTracker framework~\cite{krafka2016eye} takes the left eye, right eye, detected facial region and face location in the original frame as a binary mask (all of the size $224 \times 224$) and predicts the distance from the camera (in cm). The model is jointly trained with Euclidean loss on the x and y gaze position. The overview of the framework is shown in the second row third column image in Fig.~\ref{fig:pipelines}. 

\noindent \textbf{Multi-Branch Design.} Similar to iTracker, Jyoti et al.~\cite{jyoti2018automatic} propose a framework which takes the full face, left and right eye, both eye patch as input for inferring gaze (Refer Fig.~\ref{fig:pipelines} second-row left image). To train this network, mean squared error between the true and predicted gaze point/direction is used. 

\noindent \textbf{Two-Stream VGG Network.} In~\cite{FischerECCV2018}, a two-stream VGG network is used for gaze inference while taking left and right eye patch as input. Similar to prior works, it utilizes the sum of the individual $\ell_2$ losses between the predicted and ground truth gaze vectors to train the ensemble network.

\subsubsection{Temporal Gaze Modelling}The human gaze is a continuous and dynamic process. While scanning the environment, the concerned subject performs eye movements in terms of fixations, saccades, smooth pursuit, vergence, and vestibulo-ocular movements. Moreover, a certain image frame in time has a high correlation with the gaze direction of previous time steps. Based on this line of reasoning, several works~\cite{palmero2020benefits,gaze360_2019,zhou2019learning,wang2019neuro,palmero2018recurrent,Nonaka_2022_CVPR} have leveraged temporal information and eye movement dynamics to enhance gaze estimation performance as compared to image based static methods. Given a sequence of frames, here the task is to estimate the gaze direction of the concerned person. For this modelling, popular recurrent neural network structures have been explored (e.g. GRU~\cite{Park2020ECCV}, LSTM/bi-LSTM~\cite{gaze360_2019}).

\noindent \textbf{Multi-Modal Recurrent CNN.} Palmero et al.~\cite{palmero2018recurrent} have proposed a multimodal recurrent CNN framework in which the learned static features of all the input frames of a given sequence are fed to a many-to-one recurrent module for predicting the 3D gaze direction of the last frame in the sequence. Their approach improves the state-of-the-art gaze estimation performance significantly (i.e. by 4\% on EYEDIAP dataset). 

\noindent \textbf{DGTN.} Wang et
al.~\cite{wang2019neuro} have proposed Dynamic Gaze Transition Network (DGTN) based on semi-Markov approach which models human eye movement dynamics. DGTN first computes per-frame gaze using a CNN which is further refined using the learned dynamic information.

\noindent \textbf{Improved iTracker + bi-LSTM.} Bidirectional recurrent module based temporal modeling methods have been introduced in~\cite{zhou2019learning} which rely on both past and future frames. It is quite beneficial for low-to-mid resolution images and videos having a low frame rate ($\sim30$ fps) despite its reduced applicability for real-time application as future frames are not usually available.

\noindent \textbf{Pinball LSTM.} 
Similarly to encode contextual information along temporal domain, pinball LSTM~\cite{gaze360_2019} is proposed. This video based gaze estimation model using bidirectional LSTM considers 7-frame sequence to estimate the gaze direction of the central frame. Fig.~\ref{fig:pipelines} first row third column image illustrates the architecture of the model. The facial region from each frame is provided as input to the backbone CNN having ResNet-18 architecture. It maps input image to 256-dimensional feature space. Further, a two layer bidirectional LSTMs map these features to label space via a FC layer with an error quantile estimation i.e. ($\theta, \phi, \sigma$), where ($\theta, \phi$) is the predicted gaze direction in spherical coordinates corresponding to the ground-truth gaze vector in the eye coordinate system $g$ as $\theta = - arctan\ \frac{g_x}{g_z}$ and $\phi = arcsin\ g_y$. On the other hand, $\sigma$ corresponds to the offset from the predicted gaze i.e. $\theta + \sigma$ and $\phi + \sigma$ in the 90\% quantiles of its distribution and $\theta - \sigma$ and $\phi - \sigma$ are in the 10\% quantiles.
The pinball loss is computed as follows: given the ground truth label $y = (\theta_{gt}, \phi_{gt})$, the loss $L_{\tau}$ for the quantile $\tau$ and the angle $\alpha \in \{ \theta, \phi \}$ can be written as:

$$L_{\tau} (\alpha, \sigma, \alpha_{gt}) = max(\tau \hat{q_{\tau}}, - (1 - \tau)\hat{q_{\tau}})$$
where, $\hat{q_{\tau}} = \alpha_{gt} - (\alpha - \sigma)$, for $\tau \le 0.5$ and $\alpha_{gt} - (\alpha + \sigma)$ otherwise. This loss enforces $\theta$ and $\phi$ to converge to their ground truth
values.

\noindent \textbf{Static + LSTM.}  In an interesting study, Palmero et al.~\cite{palmero2020benefits} analyze the effect of sequential information for appearance-based gaze estimation using a static CNN network followed by a recurrent module to capture eye movement dynamics. The model is developed based on high-resolution eye-image sequences performing a stimulus-elicited fixation and saccade task in a VR scenario. The proposed model learns eye movement dynamics with accurate localization of gaze movement transition. 

\noindent \textbf{Discussion.} Despite the initial efforts that confirm the benefits of leveraging temporal information~\cite{palmero2020benefits,gaze360_2019,zhou2019learning,wang2019neuro,palmero2018recurrent}, there still have scope to explore eye dynamics in a task-driven real environment. Sometimes it is difficult to capture eye movement dynamics accurately using videos having low-resolution image frames with poor frame rates. Thus, it is still challenging how and why temporal information enhances gaze estimation performance for eye movement dynamics. Moreover, a deep understanding of eye movement patterns is required as the existing datasets are only based on task-based elicitation. It is also important to integrate the existing bio-mechanic eye models with data to achieve robust and data-efficient eye tracking. There are several open problems in this line of research in terms of eye movement dynamics (i.e. gaze directions, velocities, and gaze trajectories) in task-dependent as well as natural behaviors.

\subsubsection{Transformer based}
Transformer models have recently gotten attention for their notable performance on a broad range of vision tasks. Similarly, in the gaze estimation domain, there are two types of transformers used to date which are designed on top of the ViT framework. The first one is the \textit{pure transformer in gaze estimation (GazeTR-Pure)}~\cite{cheng2021gaze} and the other one is \textit{hybrid transformer in gaze estimation (GazeTR-Hybrid)}~\cite{cheng2021gaze}. \textit{GazeTR-Pure}~\cite{cheng2021gaze} takes the cropped face as input along with an extra token. The extra token is a learnable embedding which aggregates the image features together. On the other hand, \textit{GazeTR-Hybrid}~\cite{cheng2021gaze} is comprised of CNN and transformer. It is based on the fact that gaze estimation is a regression task and it is quite difficult to get the perception of gaze with only local patch based correlation. These models take advantage of the transformer's attention mechanism to improve gaze estimation performance. These are initial explorations using the transformer backbone. There is an immense possibility to explore this architecture.

\subsubsection{VAE/GAN based}
Variational autoencoders and GANs have been used for unsupervised or self-supervised representation learning (Refer Fig.~\ref{fig:pipelines}). Here, the latent space feature of the autoencoder model is used for gaze estimation inference~\cite{park2019few,yu2019unsupervised,zheng2020self}. Apart from representation learning, VAE and GAN based models are widely used for gaze redirection tasks~\cite{shrivastava2017learning,zheng2020self,chen2021coarse,chen2020mggr}.

\noindent \textbf{DT-ED.} For representation learning via gaze redirection in a person independent manner, variational autoencoders are often utilized~\cite{park2019few,zheng2020self}. Disentangling Transforming Encoder-Decoder (DT-ED) framework~\cite{park2019few} takes an input image $x$ and maps it to the latent space $z$ via an encoder $E$ (i.e. $E(x): x \to z$). In the latent space, DT-ED disentangles three important factors relevant to gaze, i.e. gaze direction ($z_g$), head orientation ($z_h$), and the appearance of the eye region ($z_a$). Thus, $z$ can be expressed as: $z = \{ z_a; z_g; z_h\}$. The framework disentangles these factors by explicitly applying constraints related to gaze and headpose rotations. Further, a decoder $D$ maps $z$ back to the redirected image (i.e. $ D(E(x)) : z \to \hat{x}$). The gaze direction is estimated from the $z_g$ part of the latent embedding. The overall illustration is shown in Fig.~\ref{fig:pipelines} (bottom row middle image).

\noindent \textbf{ST-ED.} Similarly, the Self-Transforming Encoder-Decoder (ST-ED) architecture~\cite{zheng2020self} (Refer Fig.~\ref{fig:pipelines} bottom row right image) takes a pair of images $x_i$ and $x_t$ as input, disentangles the subject's personal non-varying embeddings ($z^0_i$ and $z^0_t$), considers pseudo-label conditions ($\hat{c_i}$ and $\hat{c_t}$) and embedding representations ($z_i$ and $z_t$). The learning objective for transformation depends on the pseudo condition labels which consider extraneous factors in absence of ground truth annotation.

\noindent \textbf{Gaze Redirection Network.} The main motivation behind the unsupervised gaze redirection network~\cite{yu2019unsupervised} is capturing generic eye representation via gaze redirection (Refer Fig.~\ref{fig:pipelines} bottom row left image). The framework takes eye patch $I_i$ as input and predict the redirected eye patch as output $I_o$ while preserving the difference in rotation $\Delta_r = r_i - r_o = G_{\phi}(I_i) - G_{\phi}(I_o)$. In this work, gaze redirection is used as a pretext task for representation learning.

\noindent \textbf{RITnet.}
RITnet~\cite{chaudhary2019ritnet} (Refer Fig.~\ref{fig:pipelines} top row right image) is a hybrid version of U-Net and DenseNet based upon Fully Convolutional Networks (FCN). To balance the trade-off between the performance and computational complexity, it consists of 5 Down-Blocks in the encoder and 4 Up-Blocks in the decoder where the last layer of the encoder block is termed as the bottleneck layer. Each Down-Block has 5 convolution layers with LeakyReLU activation and the layers share connections with previous layers similar to DenseNet architecture. Similarly, each Up-Block has 4 convolution layers with LeakyReLU activation. All Up-Blocks have skip connection with their corresponding Down-Block which is an effective strategy to learn representation. To train the model, the following loss functions are used: \textit{1) Standard cross-entropy loss (CEL)} is applied pixel-wise to categorize each pixel into four categories (i.e. background, iris, sclera, and pupil). \textit{2) Generalized Dice Loss (GDL)} penalize the pixels on the basis of the overlap between the ground truth pixel and corresponding prediction. \textit{3) Boundary Aware Loss (BAL)} weights each pixel in terms of distance with its two nearest neighbour. This loss helps to avoid CEL confusion in boundary region. \textit{4) Surface Loss (SL)} helps to recover small regions and contours via distance based scaling. The overall loss is defined as follows:
$$ \ell = \ell_{CEL}(\lambda_1 + \lambda_2 \ell_{BAL}) + \lambda_3 \ell_{GDL} + \lambda_4 \ell_{SL}.$$

Similarly, another lightweight model~\cite{kim2019eye} uses MobileNet with depth-wise separable convolution for efficiency. It also utilize a squeeze and excitation (SE) module for performance enhancement by modelling channel independence. Moreover, the heuristic filtering of the connected component is utilized to enforce biological coherence in the network.  Few works~\cite{rot2018deep,luo2019ibug} also use multi-class classification strategy for rich representation learning.

\noindent \textbf{Other Statistical Modelling.} The statistical inference based mapping is performed based on k-NN~\cite{huang2015tabletgaze}, support vector regression~\cite{smith2013gaze,huang2015tabletgaze} and random forest~\cite{huang2015tabletgaze,sugano2014learning}. A brief overview of these methods is summarised in Table \ref{tab:gazemethod_comparison}. Prior deep learning works on semantic eye segmentation is mainly focused on iris or sclera segmentation via Fuzzy C Means clustering, Otsu’s binarization, k-NN~\cite{das2016ssrbc} etc. Sclera segmentation challenge was organized since 2015 to promote development in this area~\cite{das2016ssrbc,das2017sserbc,das2019sclera}. Recently, the OpenEDS challenge was organized in 2019 by Facebook Research in which eye segmentation was one of the sub-challenges. Most of the methods in this challenge use deep learning techniques~\cite{chaudhary2019ritnet,boutros2019eye,perry2019minenet}.

\begin{table*}[!htbp]
\caption{A comparison of gaze analysis methods with respect to registration (Reg.), representation (Represent.), Level of Supervision, Model, Prediction, validation on benchmark datasets (validation), Platforms (Plat.), Publication venue (Publ.) and year. Here, GV: Gaze Vector, Scr.: Screen, LOSO: Leave One Subject Out, LPIPS: Learned Perceptual Image Patch Similarity, MM: Morphable Model, RRF: Random Regression Forest, AEM: Anatomic Eye Model, GRN: Gaze Regression Network, ET: External Target, FV: Free Viewing, HH: HandHeld Device, HMD: Head Mounted Device, Seg.: Segmentation and GR: Gaze Redirection, LAEO: Looking At Each Other.}
\label{tab:gazemethod_comparison}
\centering
\begin{tabular}{l|c|c|l|c|c|c|c|c|c}
\toprule[0.4mm]
\rowcolor{mygray}
\multicolumn{1}{c|}{\textbf{Ref.}} & \multicolumn{1}{c|}{\textbf{Reg.}} & \multicolumn{1}{c|}{\textbf{Represent.}} & \multicolumn{1}{c|}{\begin{tabular}[c]{@{}l@{}}\textbf{Level of}\\\textbf{Sup.}\end{tabular}} & 
\multicolumn{1}{c|}{\textbf{Model}} & 
\multicolumn{1}{c|}{\textbf{Prediction}} &
\multicolumn{1}{c|}{\textbf{Validation}} &
\multicolumn{1}{c|}{\textbf{Plat.}} &
\multicolumn{1}{c|}{\textbf{Publ.}} & 
\multicolumn{1}{c}{\textbf{Year}} \\ \hline \hline
\cite{smith2013gaze} &  \begin{tabular}[c]{@{}l@{}}Face\cite{Omron} \end{tabular} &  \begin{tabular}[c]{@{}l@{}} Appear.\end{tabular} & Fully-Sup. & SVM &  \begin{tabular}[c]{@{}l@{}}Gaze locking\end{tabular}& \begin{tabular}[c]{@{}l@{}}\cite{smith2013gaze}\end{tabular} & Scr. & \begin{tabular}[c]{@{}l@{}} UIST \end{tabular}  & 2013 \\ \hline 
\cite{funes2014eyediap}  & 3-D MM  & \begin{tabular}[c]{@{}l@{}} Appear.\end{tabular} &\begin{tabular}[c]{@{}l@{}}Fully-Sup. \end{tabular}& Convex Hull & 3-D GV & \begin{tabular}[c]{@{}l@{}}~\cite{funes2014eyediap}\end{tabular} &  ET.  &\begin{tabular}[c]{@{}l@{}} ETRA \end{tabular}&  2014\\ \hline
\cite{sugano2014learning} & \begin{tabular}[c]{@{}l@{}}Face, Eye~\cite{sugano2014learning}\end{tabular}  & \begin{tabular}[c]{@{}l@{}} Appear.\end{tabular} & \begin{tabular}[c]{@{}l@{}}Fully-Sup.\end{tabular} & \begin{tabular}[c]{@{}l@{}}RRF\end{tabular} &  3-D GV & \begin{tabular}[c]{@{}l@{}}~\cite{sugano2014learning}\end{tabular} & Any & \begin{tabular}[c]{@{}l@{}}CVPR \end{tabular}  & 2014\\ \hline
\cite{wood2015rendering} & \begin{tabular}[c]{@{}l@{}}Eye\end{tabular}  & \begin{tabular}[c]{@{}l@{}}Appear. \end{tabular} & \begin{tabular}[c]{@{}l@{}} Fully-Sup.\end{tabular} &\begin{tabular}[c]{@{}l@{}} CNN+CLNF\end{tabular}& \begin{tabular}[c]{@{}l@{}}3-D GV\end{tabular}& \begin{tabular}[c]{@{}l@{}}~\cite{zhang2015appearance}\end{tabular} & Any & \begin{tabular}[c]{@{}l@{}} ICCV\end{tabular}  & 2015\\ \hline
\cite{zhang2015appearance}& \begin{tabular}[c]{@{}l@{}}Face, L/R Eye\end{tabular}  & \begin{tabular}[c]{@{}l@{}}Appear. \end{tabular} &  \begin{tabular}[c]{@{}l@{}}Fully-Sup.\end{tabular} & \begin{tabular}[c]{@{}l@{}}CNN~\cite{zhang2015appearance}\end{tabular} &  \begin{tabular}[c]{@{}l@{}}3-D GV\end{tabular} & \begin{tabular}[c]{@{}l@{}}~\cite{zhang2015appearance} \end{tabular} & Scr. & \begin{tabular}[c]{@{}l@{}}CVPR \end{tabular} &  2015\\ \hline
\cite{krafka2016eye}& \begin{tabular}[c]{@{}l@{}} Face, L/R Eye\end{tabular}  & \begin{tabular}[c]{@{}l@{}}Appear.\end{tabular} &  \begin{tabular}[c]{@{}l@{}}Fully-Sup.\end{tabular} & \begin{tabular}[c]{@{}l@{}}iTracker~\cite{krafka2016eye}\end{tabular} &  \begin{tabular}[c]{@{}l@{}}2-D Scr.\end{tabular} & \begin{tabular}[c]{@{}l@{}}\cite{huang2015tabletgaze,krafka2016eye}  \end{tabular} & HH  & \begin{tabular}[c]{@{}l@{}}CVPR \end{tabular} &  2016\\ \hline
\cite{ganin2016deepwarp}& \begin{tabular}[c]{@{}l@{}} Eye\end{tabular}  & \begin{tabular}[c]{@{}l@{}}Appear.\end{tabular} &  \begin{tabular}[c]{@{}l@{}}Fully-Sup.\end{tabular} & \begin{tabular}[c]{@{}l@{}}CNN~\cite{krafka2016eye}\end{tabular} &  \begin{tabular}[c]{@{}l@{}}GR Img.\end{tabular} & \begin{tabular}[c]{@{}l@{}}\cite{ganin2016deepwarp}  \end{tabular} & Any & \begin{tabular}[c]{@{}l@{}}ECCV \end{tabular} &  2016\\ \hline
\cite{huang2015tabletgaze}& \begin{tabular}[c]{@{}l@{}}Eye\cite{YuS} \end{tabular} & \begin{tabular}[c]{@{}l@{}}Appear. \end{tabular} & \begin{tabular}[c]{@{}l@{}}Fully-Sup.\end{tabular} & \begin{tabular}[c]{@{}l@{}}SVR \end{tabular}& 2-D Scr.& \begin{tabular}[c]{@{}l@{}}\cite{huang2015tabletgaze} \end{tabular} & HH & \begin{tabular}[c]{@{}l@{}}MVA \end{tabular} &2017\\ \hline
\cite{FischerECCV2018}& \begin{tabular}[c]{@{}l@{}}Eye~\cite{xiang2017joint}\end{tabular}  &\begin{tabular}[c]{@{}l@{}}Appear.\end{tabular} & \begin{tabular}[c]{@{}l@{}} Fully-Sup. \end{tabular} & \begin{tabular}[c]{@{}l@{}} VGG-16+FC~\cite{FischerECCV2018}\end{tabular} & \begin{tabular}[c]{@{}l@{}} 3-D GV\end{tabular} & \begin{tabular}[c]{@{}l@{}}\cite{zhang2015appearance,FischerECCV2018} \end{tabular} & Scr. & \begin{tabular}[c]{@{}l@{}}ECCV \end{tabular}  & 2018\\ \hline
\cite{park2018deep}& \begin{tabular}[c]{@{}l@{}} Eyes\end{tabular} &\begin{tabular}[c]{@{}l@{}}Appear.\end{tabular} & \begin{tabular}[c]{@{}l@{}}Fully-Sup.\end{tabular} & \begin{tabular}[c]{@{}l@{}}CNN\end{tabular}& \begin{tabular}[c]{@{}l@{}}3-D GV\end{tabular}& \begin{tabular}[c]{@{}l@{}}\cite{zhang2015appearance,funes2014eyediap}\end{tabular} & Scr. & \begin{tabular}[c]{@{}l@{}}ECCV \end{tabular}  & 2018\\ \hline
\cite{jyoti2018automatic}& \begin{tabular}[c]{@{}l@{}} Face~\cite{baltruvsaitis2016openface} \end{tabular} &\begin{tabular}[c]{@{}l@{}}Geo.+Appear.\end{tabular} & \begin{tabular}[c]{@{}l@{}}Fully-Sup.\end{tabular} & \begin{tabular}[c]{@{}l@{}}CNN~\cite{jyoti2018automatic} \end{tabular}& \begin{tabular}[c]{@{}l@{}}3-D GV\end{tabular}& \begin{tabular}[c]{@{}l@{}}\cite{huang2015tabletgaze,smith2013gaze}\end{tabular} & Desk. &  \begin{tabular}[c]{@{}l@{}} ICPR\end{tabular}  & 2018\\ \hline 
\cite{wang2018hierarchical}& \begin{tabular}[c]{@{}l@{}}Eye\end{tabular} &\begin{tabular}[c]{@{}l@{}}Geo.+Appear.\end{tabular} & \begin{tabular}[c]{@{}l@{}}Fully-Sup.\end{tabular} & \begin{tabular}[c]{@{}l@{}}HGSM+c-BiGAN\end{tabular}& \begin{tabular}[c]{@{}l@{}} Eye, GV\end{tabular}& \begin{tabular}[c]{@{}l@{}}~\cite{zhang2015appearance,funes2014eyediap} \end{tabular} &Any  & \begin{tabular}[c]{@{}l@{}}CVPR\end{tabular}  & 2018\\ \hline
\cite{chen2018appearance}& \begin{tabular}[c]{@{}l@{}}Face, L/R Eye\end{tabular} &\begin{tabular}[c]{@{}l@{}}Appear.\end{tabular} & \begin{tabular}[c]{@{}l@{}}Fully-Sup.\end{tabular} & \begin{tabular}[c]{@{}l@{}}Dilated CNN\end{tabular}& \begin{tabular}[c]{@{}l@{}} 3-D GV\end{tabular}& \begin{tabular}[c]{@{}l@{}}~\cite{zhang2015appearance,krafka2016eye,smith2013gaze} \end{tabular} & Scr. & \begin{tabular}[c]{@{}l@{}}ACCV\end{tabular}  & 2018\\ \hline
\cite{park2019few}& \begin{tabular}[c]{@{}l@{}}Face \end{tabular} &\begin{tabular}[c]{@{}l@{}} Appear.\end{tabular} & \begin{tabular}[c]{@{}l@{}}Few-Shot\end{tabular} & \begin{tabular}[c]{@{}l@{}}DT-ED+ML \end{tabular}& \begin{tabular}[c]{@{}l@{}}3-D GV\end{tabular}& \begin{tabular}[c]{@{}l@{}}~\cite{zhang2015appearance,krafka2016eye}\end{tabular} & Scr. &  \begin{tabular}[c]{@{}l@{}} ICCV\end{tabular}  & 2019\\ \hline
\cite{gaze360_2019}& \begin{tabular}[c]{@{}l@{}}Face\end{tabular} &\begin{tabular}[c]{@{}l@{}} Appear.\end{tabular} & \begin{tabular}[c]{@{}l@{}}Fully-Sup.\end{tabular} & \begin{tabular}[c]{@{}l@{}}Pinball LSTM \end{tabular}& \begin{tabular}[c]{@{}l@{}}3-D GV\end{tabular}& \begin{tabular}[c]{@{}l@{}}~\cite{zhang2015appearance,smith2013gaze,huang2015tabletgaze}\end{tabular} & ET &  \begin{tabular}[c]{@{}l@{}}ICCV\end{tabular}  & 2019\\ \hline
\begin{tabular}[c]{@{}l@{}}\cite{garbin2019openeds}\end{tabular}& Eye  & Appear. &\begin{tabular}[c]{@{}l@{}} Fully-Sup.\end{tabular} & SegNet~\cite{badrinarayanan2017segnet} &  Seg. Map &\begin{tabular}[c]{@{}l@{}}~\cite{garbin2019openeds} \end{tabular}  &  HMD & \begin{tabular}[c]{@{}l@{}}ICCVW \end{tabular}&2019\\ \hline
\cite{wang2019neuro}& \begin{tabular}[c]{@{}l@{}}Face, L/R Eye\end{tabular} &\begin{tabular}[c]{@{}l@{}}Appear.\end{tabular} & \begin{tabular}[c]{@{}l@{}}Fully-Sup.\end{tabular} & \begin{tabular}[c]{@{}l@{}}DGTN\end{tabular}& \begin{tabular}[c]{@{}l@{}}GV\end{tabular}& \begin{tabular}[c]{@{}l@{}}~\cite{wang2019neuro}\end{tabular} & Desk.  & \begin{tabular}[c]{@{}l@{}}CVPR\end{tabular}  & 2019\\ \hline
\cite{xiong2019mixed}& \begin{tabular}[c]{@{}l@{}}Face\end{tabular} &\begin{tabular}[c]{@{}l@{}}Appear.\end{tabular} & \begin{tabular}[c]{@{}l@{}}Fully-Sup.\end{tabular} & \begin{tabular}[c]{@{}l@{}}MeNet\end{tabular}& \begin{tabular}[c]{@{}l@{}}3-D GV\end{tabular}& \begin{tabular}[c]{@{}l@{}}~\cite{zhang2015appearance,sugano2014learning,krafka2016eye}\end{tabular} & Scr. & \begin{tabular}[c]{@{}l@{}}CVPR\end{tabular}  & 2019\\ \hline
\cite{wang2019generalizing}& \begin{tabular}[c]{@{}l@{}}Face, Eye\end{tabular} &\begin{tabular}[c]{@{}l@{}}Appear.\end{tabular} & \begin{tabular}[c]{@{}l@{}}Semi/Unsup.\end{tabular} & \begin{tabular}[c]{@{}l@{}}BCNN\end{tabular}& \begin{tabular}[c]{@{}l@{}}3-D GV\end{tabular}& \begin{tabular}[c]{@{}l@{}}~\cite{zhang2015appearance,funes2014eyediap}\end{tabular} & Desk.   &\begin{tabular}[c]{@{}l@{}}CVPR\end{tabular}  & 2019\\ \hline
\cite{chaudhary2019ritnet}& \begin{tabular}[c]{@{}l@{}}Eyes\end{tabular} &\begin{tabular}[c]{@{}l@{}}Appear.\end{tabular} & \begin{tabular}[c]{@{}l@{}}Fully-Sup.\end{tabular} & \begin{tabular}[c]{@{}l@{}}Hybrid U-net\end{tabular}& \begin{tabular}[c]{@{}l@{}}Seg. Map\end{tabular}& \begin{tabular}[c]{@{}l@{}}~\cite{garbin2019openeds} \end{tabular} & HMD   &\begin{tabular}[c]{@{}l@{}}ICCVW\end{tabular}  & 2019\\ \hline
\cite{kansal2019eyenet}& \begin{tabular}[c]{@{}l@{}}Eyes\end{tabular} &\begin{tabular}[c]{@{}l@{}}Appear.\end{tabular} & \begin{tabular}[c]{@{}l@{}}Fully-Sup.\end{tabular} & \begin{tabular}[c]{@{}l@{}}Modified Resnet\end{tabular}& \begin{tabular}[c]{@{}l@{}}Seg. Map\end{tabular}& \begin{tabular}[c]{@{}l@{}}~\cite{garbin2019openeds} \end{tabular} & HMD   &\begin{tabular}[c]{@{}l@{}}ICCVW\end{tabular}  & 2019\\ \hline
\cite{boutros2019eye}& \begin{tabular}[c]{@{}l@{}}Eyes\end{tabular} &\begin{tabular}[c]{@{}l@{}}Appear.\end{tabular} & \begin{tabular}[c]{@{}l@{}}Fully-Sup.\end{tabular} & \begin{tabular}[c]{@{}l@{}}Eye-MMS\end{tabular}& \begin{tabular}[c]{@{}l@{}}Seg. Map\end{tabular}& \begin{tabular}[c]{@{}l@{}}~\cite{garbin2019openeds} \end{tabular} & HMD  &\begin{tabular}[c]{@{}l@{}}ICCVW\end{tabular}  & 2019\\ \hline
\cite{perry2019minenet}& \begin{tabular}[c]{@{}l@{}}Eyes\end{tabular} &\begin{tabular}[c]{@{}l@{}}Appear.\end{tabular} & \begin{tabular}[c]{@{}l@{}}Fully-Sup.\end{tabular} & \begin{tabular}[c]{@{}l@{}}Dilated CNN\end{tabular}& \begin{tabular}[c]{@{}l@{}}Seg. Map\end{tabular}& \begin{tabular}[c]{@{}l@{}}~\cite{garbin2019openeds} \end{tabular} & HMD   &\begin{tabular}[c]{@{}l@{}}ICCVW\end{tabular}  & 2019\\ \hline
\cite{yu2019improving}& \begin{tabular}[c]{@{}l@{}}Eyes\end{tabular} &\begin{tabular}[c]{@{}l@{}}Appear.+Seg.\end{tabular} & \begin{tabular}[c]{@{}l@{}}Few-shot\end{tabular} & \begin{tabular}[c]{@{}l@{}}GR \end{tabular}& \begin{tabular}[c]{@{}l@{}}2-D GV\end{tabular}& \begin{tabular}[c]{@{}l@{}}\cite{zhang2015appearance,smith2013gaze} \end{tabular} & Any & \begin{tabular}[c]{@{}l@{}}CVPR\end{tabular}  & 2019\\ \hline 
\cite{yu2019unsupervised}& \begin{tabular}[c]{@{}l@{}}Eyes\end{tabular} &\begin{tabular}[c]{@{}l@{}}Appear.\end{tabular} & \begin{tabular}[c]{@{}l@{}}Unsup.\end{tabular} & \begin{tabular}[c]{@{}l@{}}GR\end{tabular}& \begin{tabular}[c]{@{}l@{}}2-D GV\end{tabular}& \begin{tabular}[c]{@{}l@{}}\cite{zhang2015appearance,smith2013gaze} \end{tabular}   & Any & \begin{tabular}[c]{@{}l@{}}CVPR\end{tabular}  & 2019\\ \hline
\cite{dubey2019unsupervised}& \begin{tabular}[c]{@{}l@{}}Face, Eye \end{tabular} &\begin{tabular}[c]{@{}l@{}} Appear.\end{tabular} & \begin{tabular}[c]{@{}l@{}}Unsup.\end{tabular} & \begin{tabular}[c]{@{}l@{}}IzeNet\end{tabular}& \begin{tabular}[c]{@{}l@{}}3-D GV\end{tabular}& \begin{tabular}[c]{@{}l@{}}\cite{smith2013gaze,huang2015tabletgaze} \end{tabular} & FV & \begin{tabular}[c]{@{}l@{}}IJCNN\end{tabular}  & 2019\\ \hline
\cite{buhler2019content}& \begin{tabular}[c]{@{}l@{}}Eye\end{tabular} &\begin{tabular}[c]{@{}l@{}}Appear.+Seg.\end{tabular} & \begin{tabular}[c]{@{}l@{}}Fully-Sup.\end{tabular} & \begin{tabular}[c]{@{}l@{}}Seg2Eye\end{tabular}& \begin{tabular}[c]{@{}l@{}}Eye Img.\end{tabular}& \begin{tabular}[c]{@{}l@{}}~\cite{garbin2019openeds}\end{tabular} & HMD & \begin{tabular}[c]{@{}l@{}}ICCVW\end{tabular}  & 2019\\ \hline
\cite{zhu2020hierarchical}& \begin{tabular}[c]{@{}l@{}}Eye Seq.\end{tabular} &\begin{tabular}[c]{@{}l@{}} Appear.\end{tabular} & \begin{tabular}[c]{@{}l@{}}Unsup.\end{tabular} & \begin{tabular}[c]{@{}l@{}}Hier. HMM\end{tabular}& \begin{tabular}[c]{@{}l@{}}Eye Move.\end{tabular}& \begin{tabular}[c]{@{}l@{}}~\cite{komogortsev2013automated} \end{tabular} & Any & \begin{tabular}[c]{@{}l@{}}ECCVW\end{tabular}  & 2019\\ \hline
\cite{shen2020domain}& \begin{tabular}[c]{@{}l@{}}Eye\end{tabular} &\begin{tabular}[c]{@{}l@{}}Appear.\end{tabular} & \begin{tabular}[c]{@{}l@{}}Semi/Unsup.\end{tabular} & \begin{tabular}[c]{@{}l@{}}mSegNet+Discre.\end{tabular}& \begin{tabular}[c]{@{}l@{}}Seg. Map\end{tabular}& \begin{tabular}[c]{@{}l@{}}~\cite{garbin2019openeds} \end{tabular} & HMD & \begin{tabular}[c]{@{}l@{}}ECCVW\end{tabular}  & 2019\\ \hline
\cite{perry2020eyeseg}& \begin{tabular}[c]{@{}l@{}}Eye\end{tabular} &\begin{tabular}[c]{@{}l@{}}Appear.\end{tabular} & \begin{tabular}[c]{@{}l@{}}Few-Shot\end{tabular} & \begin{tabular}[c]{@{}l@{}}EyeSeg\end{tabular}& \begin{tabular}[c]{@{}l@{}}Seg. Map\end{tabular}& \begin{tabular}[c]{@{}l@{}}~\cite{garbin2019openeds} \end{tabular} & HMD  & \begin{tabular}[c]{@{}l@{}}ECCVW\end{tabular}  & 2019\\ \hline
\cite{zheng2020self}& \begin{tabular}[c]{@{}l@{}}Face\end{tabular} &\begin{tabular}[c]{@{}l@{}}Appear.\\ \end{tabular} & \begin{tabular}[c]{@{}l@{}}Fully-Sup.\end{tabular} & \begin{tabular}[c]{@{}l@{}}ST-ED\end{tabular}& \begin{tabular}[c]{@{}l@{}}GR\end{tabular}& \begin{tabular}[c]{@{}l@{}}\cite{krafka2016eye,smith2013gaze,funes2014eyediap}\end{tabular} &Scr. & \begin{tabular}[c]{@{}l@{}}NeurIPS\end{tabular}  & 2020\\ \hline 
\begin{tabular}[c]{@{}l@{}}\cite{palmero2020openeds2020} \end{tabular} & Eye  & Appear.   & \begin{tabular}[c]{@{}l@{}}Fully-Sup.\end{tabular} & Modified ResNet & GR Img.   & \begin{tabular}[c]{@{}l@{}}\cite{palmero2020openeds2020}\end{tabular}    &  HMD & \begin{tabular}[c]{@{}l@{}}ECCVW \end{tabular}&2020 \\ \hline
\cite{Park2020ECCV}& \begin{tabular}[c]{@{}l@{}}Eyes\end{tabular}& \begin{tabular}[c]{@{}l@{}}Appear.\end{tabular} & \begin{tabular}[c]{@{}l@{}}Fully-Sup.\end{tabular} & \begin{tabular}[c]{@{}l@{}}ResNet-18+GRU\end{tabular}& \begin{tabular}[c]{@{}l@{}}PoG,3-D GV\end{tabular}& \begin{tabular}[c]{@{}l@{}}\cite{Park2020ECCV} \end{tabular} &Scr.& \begin{tabular}[c]{@{}l@{}}ECCV\end{tabular}  & 2020\\ \hline
\cite{zhang2020eth} & \begin{tabular}[c]{@{}l@{}}Face\end{tabular}  & \begin{tabular}[c]{@{}l@{}}Appear.\end{tabular} & \begin{tabular}[c]{@{}l@{}}Fully-Sup.\end{tabular} & ResNet-50 & 3-D GV & \begin{tabular}[c]{@{}l@{}}~\cite{zhang2015appearance,krafka2016eye,gaze360_2019,funes2014eyediap}\end{tabular}  &Scr. & \begin{tabular}[c]{@{}l@{}}ECCV \end{tabular} &2020\\\hline
\cite{dias2020gaze}& \begin{tabular}[c]{@{}l@{}}Face\end{tabular} &\begin{tabular}[c]{@{}l@{}}Appear.\end{tabular} & \begin{tabular}[c]{@{}l@{}}Semi-Sup.\end{tabular} & \begin{tabular}[c]{@{}l@{}}GRN\end{tabular}& \begin{tabular}[c]{@{}l@{}}GV\end{tabular}& \begin{tabular}[c]{@{}l@{}}~\cite{NIPS2015_ec895663}\end{tabular} & FV  &\begin{tabular}[c]{@{}l@{}}WACV\end{tabular}  & 2020\\\hline
\cite{zhang2020learning}& \begin{tabular}[c]{@{}l@{}}Face, Eye\end{tabular} &\begin{tabular}[c]{@{}l@{}}Appear.\end{tabular} & \begin{tabular}[c]{@{}l@{}}Fully-Sup.\end{tabular} & \begin{tabular}[c]{@{}l@{}}RSN+GazeNet\end{tabular}& \begin{tabular}[c]{@{}l@{}}GV\end{tabular}& \begin{tabular}[c]{@{}l@{}}\cite{funes2014eyediap,zhang2015appearance,krafka2016eye} \end{tabular} & Scr. &\begin{tabular}[c]{@{}l@{}}BMVC\end{tabular}  & 2020\\\hline
\cite{cheng2020coarse}& \begin{tabular}[c]{@{}l@{}}Face, Eye\end{tabular} &\begin{tabular}[c]{@{}l@{}}Appear.\end{tabular} & \begin{tabular}[c]{@{}l@{}}Fully-Sup.\end{tabular} & \begin{tabular}[c]{@{}l@{}}CA-Net\end{tabular}& \begin{tabular}[c]{@{}l@{}}GV\end{tabular}& \begin{tabular}[c]{@{}l@{}}~\cite{zhang2015appearance,funes2014eyediap} \end{tabular} &  Scr. &\begin{tabular}[c]{@{}l@{}}AAAI\end{tabular}  & 2020\\\hline
\cite{cheng2020gaze}& \begin{tabular}[c]{@{}l@{}}Face, Eye\end{tabular} &\begin{tabular}[c]{@{}l@{}}Appear.\end{tabular} & \begin{tabular}[c]{@{}l@{}}Fully-Sup.\end{tabular} & \begin{tabular}[c]{@{}l@{}} FAR-Net\end{tabular}& \begin{tabular}[c]{@{}l@{}}GV\end{tabular}& \begin{tabular}[c]{@{}l@{}}\cite{funes2014eyediap,zhang2015appearance,FischerECCV2018} \end{tabular} & Scr.  &\begin{tabular}[c]{@{}l@{}}TIP\end{tabular}  & 2020\\\hline
\cite{chen2021coarse}& \begin{tabular}[c]{@{}l@{}}Eye\end{tabular} &\begin{tabular}[c]{@{}l@{}}Appear.+AEM\end{tabular} & \begin{tabular}[c]{@{}l@{}}Fully-Sup.\end{tabular} & \begin{tabular}[c]{@{}l@{}}MT c-GAN\end{tabular}& \begin{tabular}[c]{@{}l@{}} Eye Img.\end{tabular}& \begin{tabular}[c]{@{}l@{}}~\cite{zhang2015appearance,smith2013gaze,sugano2014learning} \end{tabular} & Scr. & \begin{tabular}[c]{@{}l@{}}WACV\end{tabular}  & 2021\\ \hline
\cite{bao2021adaptive}& \begin{tabular}[c]{@{}l@{}}Face, Eye\end{tabular} &\begin{tabular}[c]{@{}l@{}}Appear.\end{tabular} & \begin{tabular}[c]{@{}l@{}}Fully-Sup.\end{tabular} & \begin{tabular}[c]{@{}l@{}} AFF-Net\end{tabular}& \begin{tabular}[c]{@{}l@{}}Scr., GV\end{tabular}& \begin{tabular}[c]{@{}l@{}}~\cite{krafka2016eye,zhang2017s}\end{tabular} &  Scr.  &\begin{tabular}[c]{@{}l@{}}Arxiv\end{tabular}  & 2021\\\hline
\cite{cheng2021puregaze}& \begin{tabular}[c]{@{}l@{}}Face\end{tabular} &\begin{tabular}[c]{@{}l@{}}Appear.\end{tabular} & \begin{tabular}[c]{@{}l@{}}Unsup.\end{tabular} & \begin{tabular}[c]{@{}l@{}}PureGaze\end{tabular}& \begin{tabular}[c]{@{}l@{}}Face, GV\end{tabular}& \begin{tabular}[c]{@{}l@{}}~\cite{zhang2020eth,gaze360_2019,zhang2015appearance,sugano2014learning}\end{tabular} &  Scr.  &\begin{tabular}[c]{@{}l@{}}Arxiv\end{tabular}  & 2021\\ \hline 
\cite{Kothari_2021_CVPR}& \begin{tabular}[c]{@{}l@{}}Face\end{tabular} &\begin{tabular}[c]{@{}l@{}}Appear.\end{tabular} & \begin{tabular}[c]{@{}l@{}}Weakly-Sup.\end{tabular} & \begin{tabular}[c]{@{}l@{}}ResNet-18+LSTM\end{tabular}& \begin{tabular}[c]{@{}l@{}}GV\end{tabular}& \begin{tabular}[c]{@{}l@{}}~\cite{zhang2020eth,gaze360_2019,Kothari_2021_CVPR,krafka2016eye}\end{tabular} &  Any  &\begin{tabular}[c]{@{}l@{}}CVPR\end{tabular}  & 2021\\ \hline
\cite{marin2021pami}& \begin{tabular}[c]{@{}l@{}}Face\end{tabular} &\begin{tabular}[c]{@{}l@{}}Appear.\end{tabular} & \begin{tabular}[c]{@{}l@{}}Fully-Sup.\end{tabular} & \begin{tabular}[c]{@{}l@{}}LAEO-Net++\end{tabular}& \begin{tabular}[c]{@{}l@{}}LAEO\end{tabular}& \begin{tabular}[c]{@{}l@{}}~\cite{marin2019laeo}\end{tabular} &  Any  &\begin{tabular}[c]{@{}l@{}}TPAMI\end{tabular}  & 2021\\ \hline
\cite{ghosh2022mtgls}& \begin{tabular}[c]{@{}l@{}}Face, Eye\end{tabular} &\begin{tabular}[c]{@{}l@{}}Appear.\end{tabular} & \begin{tabular}[c]{@{}l@{}}Limited-Sup.\end{tabular} & \begin{tabular}[c]{@{}l@{}}ResNet-50\end{tabular}& \begin{tabular}[c]{@{}l@{}}GV\end{tabular}& \begin{tabular}[c]{@{}l@{}}~\cite{smith2013gaze,zhang2015appearance,gaze360_2019,ghosh2020speak2label}\end{tabular} &  Any  &\begin{tabular}[c]{@{}l@{}}WACV\end{tabular}  & 2022\\
\bottomrule[0.4mm]
\end{tabular}
\vspace{-5mm}
\end{table*}

\subsubsection{Discussion} 
In an attempt to summarize the recent deep network based gaze analysis methods, we present some main take away points as follows:

\begin{itemize}[topsep=1pt,itemsep=0pt,partopsep=1ex,parsep=1ex,leftmargin=*]
    \item The overall gaze estimation methods are divided into two broad categories: \textit{1) 2-D Gaze Estimation:} In this context the proposed methods map the input image to 2-D Point of Regard (PoR) in the visual plane. The visual planes could either be the observable object or screen. Non deep learning methods or early deep learning methods \cite{hansen2009eye,zhang2015appearance,zhang2017mpiigaze,zhang2017s} perform these mappings. \textit{2) 3-D Gaze Estimation:} The 3-D gaze estimation basically considers the gaze vector instead of 2-D PoR. The gaze vector is the line joining the pupil center point with the point of regard. Recent works \cite{zhang2020eth,park2018deep,park2019few,Park2020ECCV,Kothari_2021_CVPR} mainly relies on 3-D gaze estimation methods. The choice of gaze estimation methods rely on the application and requirement. 
    \item Single branch CNN based architectures~\cite{zhang2015appearance,zhang2017s,zhang2017mpiigaze,krafka2016eye,park2018deep,wang2018hierarchical} are widely used over the past few years for progressive improvements on benchmark datasets. The input to these networks are restricted to single eye, eye patch or face. Thus, to further boost the performance, multi branch networks are proposed which utilize eyes, face, geometric constraints, visual plane grid as input.
    \item Both single or multi branch networks depend on spatial information. However, eye movement is dynamic in nature. Thus, few recent proposed architectures \cite{gaze360_2019,wang2018hierarchical} use temporal information for inference. 
    \item For representation learning, VAE and GAN based architectures~\cite{park2019few,yu2019unsupervised,zheng2020self} are explored. However, it is observed that these architectures could have high time complexity as compared to single or multi branch CNN. 
    \item \textit{Prior based appearance encoding} is another line of approaches for encoding rich feature representation. Few works have defined priors based on eye anatomy~\cite{park2018deep}, geometrical constraint~\cite{cheng2020gaze} as biases for better generalization. Despite direct appearance encoding, Park et al.~\cite{park2018deep} proposed an intermediate pictorial representation, termed a `gazemap' (refer Fig.~\ref{fig:pipelines}) of the eye to simplify the gaze estimation task. Similarly, the `two eye asymmetry' property is utilized for gaze estimation~\cite{cheng2020gaze} where the underlying hypothesis is that despite the difference in appearances of two eyes due to environmental factors, the gaze directions remains approximately the same. The CNN based regression model is assumed to be independent of identity distribution, however, due to the subject-specific offset of the nodal point of the eyes, gaze datasets have identity specific bias. Xiong et al.~\cite{xiong2019mixed} inject this bias as a prior by mixing different models. Similarly, to handle this offset, the gaze is decomposed into the subject independent and dependent bias for performance enhancement and better generalization~\cite{chen2020offset}.
    \item In order to train the deep learning based models, $\ell_2$~\cite{zhang2015appearance,zhang2017s,zhang2017mpiigaze} and cosine similarity based losses \cite{zhang2020eth,Park2020ECCV} are used. However, a novel pinball loss~\cite{gaze360_2019} is proposed to model the uncertainty in gaze estimation, especially in unconstrained settings. 
    \item Similarly for deep learning based eye segmentation approaches, the eye image to segmentation mapping is performed in a non-parametric way which implicitly encodes shape, geometry, appearance and other factors~\cite{luo2020shape,kansal2019eyenet,rot2018deep,chaudhary2019ritnet,wu2019eyenet,garbin2019openeds}. The most popular network architectures for eye segmentation are U-net~\cite{ronneberger2015u}, modified version of SegNet~\cite{garbin2019openeds}, RITnet~\cite{chaudhary2019ritnet}, EyeNet~\cite{kansal2019eyenet}. These VAE based architechtures have high time and space complexity. However, recent methods \cite{kansal2019eyenet,chaudhary2019ritnet} do consider these factors without compromising the performance.
\end{itemize}

\subsection{Level of Supervision} \label{sec:supervision}
Based on the type of supervision, the training procedure can be
classified into the following categories: \textit{fully-supervised, Semi-/Self-/weakly-/unsupervised}.

\subsubsection{Fully-Supervised.} Supervised learning paradigm is the most commonly used training framework in gaze estimation literature~\cite{smith2013gaze,sugano2014learning,zhang2015appearance,huang2015tabletgaze,park2018deep,park2018learning} and eye segmentation literature~\cite{chaudhary2019ritnet,kansal2019eyenet,boutros2019eye,perry2019minenet,das2013sclera,das2016ssrbc,das2017sserbc,das2019sclera,chen2019unsupervised}. As the fully-supervised methods require a lot of accurately annotated data. Accurate annotation of gaze data is a complex, noise-prone, and time-consuming task and sometimes it requires expensive data acquisition setups. Moreover, there is a high possibility of noisy or wrong annotation due to distraction in participation during data collection, eye blink activity and inherent measurement errors in data curation settings. Variation in data curation setup limits merging multiple datasets for supervision. Dataset specific data acquisition processes are discussed in Sec.~\ref{sec:datasets}. Thus, the research community is moving towards learning with less supervision.  

\noindent \textbf{Multi-Task Learning.}
Multi-task learning incorporates different tasks which provide auxiliary information as a bias to improve model performance. The auxiliary information can be Gaze+Landmark~\cite{yu2018deep}, PoG+Screen saliency~\cite{Park2020ECCV,wang2019inferring}, Gaze+Depth~\cite{lian2019rgbd}, Gaze+Headpose~\cite{zhu2017monocular}, Segmentation+Gaze~\cite{wu2019eyenet} and Gaze-direction+Gaze-uncertainty~\cite{gaze360_2019}. These gaze aligned tasks facilitate strong representation learning with additional task based supervision.
\begin{figure*}[t]
    \centering
    \includegraphics[width=\linewidth]{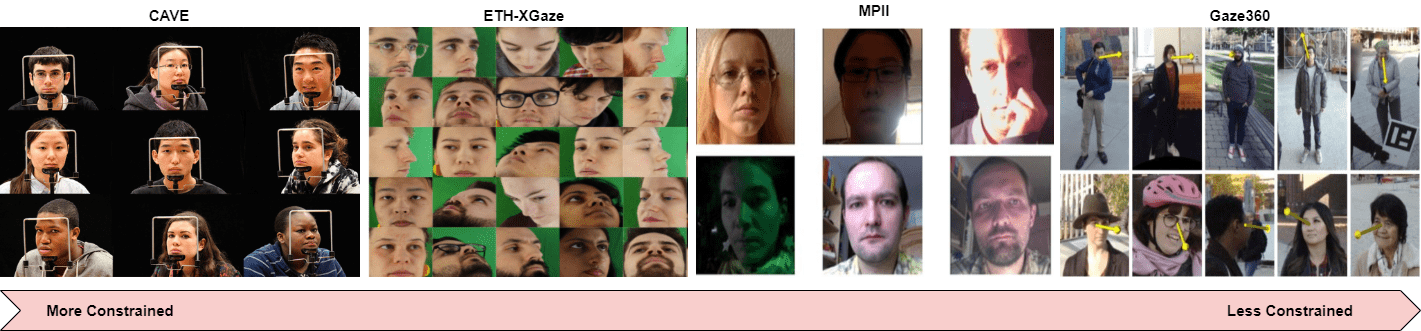}
    \caption{Data collection procedure in different settings for benchmark datasets. From left to right the examples are from CAVE~\cite{smith2013gaze}, Eth-XGaze~\cite{zhang2020eth}, MPII~\cite{zhang2017mpiigaze} and Gaze360~\cite{gaze360_2019} datasets. The leftmost one is more constrained and the rightmost one is less constrained. Images are taken from respective datasets~\cite{gaze360_2019,zhang2020eth,zhang2017mpiigaze,smith2013gaze}. Refer Table \ref{tab:gazedataset_comparison} for more details.}
    \label{fig:data_collection}
    \vspace{-5mm}
\end{figure*}

\subsubsection{Semi-/Self-/Weakly-/Unsupervised.}
To a large extent, the supervised deep learning based methods' performance depends on the quality and quantity of annotated data. However, manual labeling of gaze data is a complex, time consuming and labor extensive process. On this front, Semi-/Self-/Weakly-/Unsupervised Learning paradigms provide a promising alternative to enable learning from a vast amount of readily available non-annotated data. For learning paradigms with less supervision, the important methods are described in detail below: 

\noindent \textbf{Weakly-supervised and Learning from Pseudo Labels.}  
Weakly supervised learning aims to bridge the gap between the fully-supervised and fully-unsupervised techniques. Till date in gaze estimation domain, the weak supervision has been performed via `Looking At Each Other (LAEO)'~\cite{kothari2020gaze} and pseudo labelling~\cite{ghosh2022mtgls}. For weak supervision, Kothari et al.~\cite{kothari2020gaze} leverage strong gaze-related geometric constraints from two people interaction scenario. On the other hand, MTGLS~\cite{ghosh2022mtgls} framework leverages from non-annotated facial image data by three complementary signals i.e. (1) the line of sight of the pupil, (2) the head-pose and (3) the eye dexterity.

\noindent \textbf{Unsupervised and Self-supervised Representation Learning.} Self supervised learning has emerged as a popular technique for learning meaningful representations from vast amount of non-annotated data. It requires pseudo labels for any pre-designed task which is often termed as \textit{Auxiliary} or \textit{Pretext} Task. The pre-designed task is mostly aligned with the gaze estimation. Dubey et al.~\cite{dubey2019unsupervised} propose a pretext task where the visual regions of the gaze are divided into zones by geometric constraints. These pseudo labels are utilized for representation learning. Yu et al.~\cite{yu2019unsupervised} use subject specific gaze redirection as a pretext task. Swapping Affine Transformations (SwAT)~\cite{farkhondeh2022towards} is the extended version of Swapping Assignments Between Views (SwAV), a popular self supervised learning framework used for gaze representation learning using different augmentation techniques. The self-supervised representation learning has the potential to eliminate the major drawback of gaze data annotation which is quite difficult and error prone. Future directions in this area may include designing better pre-text tasks and combining multiple pre-text tasks to jointly pre-train the models~\cite{ghosh2022mtgls}. In addition, combining data-driven eye tracking with model-based eye tracking can be another future direction as model-based eye tracking can provide pseudo-labels or pre-train the models.

\noindent \textbf{Few Shot Learning.}
Few-shot learning aims to adapt to a new task with very few examples~\cite{park2019few,yu2019improving}. The main challenge in few shot paradigm is over-fitting issue since highly over-parameterized deep networks are involved to learn from only a few training samples. To this front, mainly gaze redirection strategy~\cite{yu2019improving} and Few-shot Adaptive GaZE Estimation (FAZE)~\cite{park2019few} frameworks are proposed. Among them, FAZE is shown a two stage adaptation strategy. In the first stage, a rotation aware latent space embedding is learned based on encoder-decoder framework. Further, adaptation is performed on top of the features using MAML which is a popular meta-learning paradigm. FAZE is able to adapt to a new subject with $\leq 9$ sample which is quite promising.


\noindent \textbf{Learning-by-synthesis.} The term `learning-by-synthesis' is coined by Sugano et al.~\cite{sugano2014learning}. The main objective is to synthesize different gaze viewpoints to multiply the data from a quantitative and a qualitative perspectives rather than manual labelling. Few other studies~\cite{sugano2014learning,ganin2016deepwarp,he2019photo,wood2018gazedirector,yu2019improving,kaur2021subject,wang2018hierarchical} also adopt data generation methods which can address the diversity in terms of headpose and eye rotation. However, these generative models have high computational complexity and are constrained by the quality of generated images.

\noindent \textbf{Discussion.} In gaze estimation domain, there are still very few works in semi-/self-/weakly-/unsupervised learning paradigms. Among these works, learning from pseudo labels have their limitations as the label space contains noise. On the other hand, gaze redirection synthesis is based on the availability of same or different person's data with eye rotation or the prior knowledge of rotation angle. Thus, learning robust and generalizable gaze representations using minimal or no supervision still remains an open-ended research question. One possible direction to alleviate this problem is to combine data-driven eye tracking with model-based eye tracking to produce physically plausible  eye tracking models that are data efficient during training and generalize better during testing.

\section{Validation} \label{ref:validation}
In this section, we review the commonly followed evaluation procedures on various datasets along with the metrics adopted in the literature.

\subsection{Datasets for Gaze Analysis}\label{sec:datasets}
With the rapid progress in the gaze analysis domain, several datasets have been proposed for different gaze analysis tasks (see Sec.~\ref{sec:task}). The dataset collection technique has evolved from constrained lab environments~\cite{smith2013gaze} to unconstrained indoor~\cite{huang2015tabletgaze,zhang2017mpiigaze,zhang2017s,FischerECCV2018} and outdoor settings~\cite{gaze360_2019} (Refer Fig.~\ref{fig:data_collection}). We provide a detailed overview of the datasets in Table \ref{tab:gazedataset_comparison}. 
Compared with early datasets~\cite{smith2013gaze,funes2014eyediap}, recently released datasets~\cite{gaze360_2019,Park2020ECCV} are typically more advanced with less bias, improved complexity, and larger in scale. These are better suited for training and evaluation. We describe a few important datasets below:

\noindent \textit{CAVE}~\cite{smith2013gaze} contains 5,880 images of 56 subjects with different gaze directions and head poses. There are 21 different gaze directions for each person and the data was collected in a constrained lab environment, with 7 horizontal and 3 vertical gaze locations. 

\noindent The \textit{Eyediap} dataset~\cite{FunesMora_ETRA_2014} was designed to overcome the main challenges associated with the head pose, person and 3-D target variations along with changes in ambient and sensing conditions. 

\noindent \textit{TabletGaze}~\cite{huang2015tabletgaze} is a large unconstrained dataset of 51 subjects with 4 different postures and 35 gaze locations collected using a tablet in an indoor environment. TabletGaze dataset is also collected in a $7\times5$ grid format. 

\noindent \textit{MPII}~\cite{zhang2017mpiigaze} gaze dataset contains 213,659 images collected from 15 subjects during natural everyday events in front of a laptop over a three-month duration. MPII gaze dataset is collected by showing random points on the laptop screen to the participants. Further, Zhang et al.~\cite{zhang2017s} curate \textit{MPIIFaceGaze} dataset with the hypothesis that gaze can be more accurately predicted when the entire face is considered. 

\noindent \textit{RT-GENE} dataset~\cite{FischerECCV2018} is recorded in a more naturalistic environment with varied gaze and head pose angles. The ground truth annotation was done using a motion capture system with mobile eye-tracking glasses. 

\noindent \textit{Gaze360}~\cite{gaze360_2019} is a large-scale gaze estimation dataset collected from 238 subjects in unconstrained indoor and outdoor settings with a wide range of head pose.

\noindent \textit{ETH-XGaze}~\cite{zhang2020eth} is a large scale dataset collected in a constraint environment with a wide range of head pose, high-resolution images. The dataset contains images from different camera positions, illumination conditions to add more challenges to the data.  

\noindent \textit{EVE}~\cite{Park2020ECCV} is also collected in constraint indoor setting with different camera views to map human gaze in screen co-ordinate. 

Similar to gaze estimation several benchmark datasets have been proposed over the past few years for eye and sclera segmentation. The datasets collected for sclera segmentation is in a constraint environment and with very few subjects~\cite{derakhshani2006new,derakhshani2007texture,crihalmeanu2009enhancement}. A more challenging publicly available dataset was released in sclera recognition challenges~\cite{das2016ssrbc,das2017sserbc,das2019sclera}. Recently, a large scale dataset termed as \textit{OpenEDS: Open Eye Dataset}~\cite{garbin2019openeds}, is released which contains eye images collected by using a VR head-mounted device. Additionally, there was two synchronized eye facing cameras having a frame rate of 200 Hz. The data was collected under controlled illumination and contains 12,759 images with eye segmentation masks collected from 152 participants.


\begin{table*}[!htbp]
\caption{\textbf{Datasets.} A comparison of gaze datasets with respect to several attributes (i.e. number of subjects (\# sub), gaze labels, modality, headpose and gaze angle in yaw and pitch axis, environment (Env.), baseline method, data statistics (\# data), and year of publication.) The abbreviations used are: In: Indoor, Out: Outdoor, Both: Indoor + Outdoor, Gen.: Generation, u/k: unknown, Seq.: Sequence, VF: Visual Field, EB: Eye Blink, GE: Gaze Event~\cite{kothari2020gaze}, GBRT: Gradient Boosting Regression Trees, GC: Gaze Communication, GNN: Graph Neural Network and Seg.: Segmentation.}
\label{tab:gazedataset_comparison}
\centering
\scalebox{0.95}{
\begin{tabular}{l|c|c|l|c|c|c|l|l|c}
\toprule[0.4mm]
\rowcolor{mygray}
\multicolumn{1}{c|}{\textbf{Dataset}} & \multicolumn{1}{c|}{\textbf{\# Sub}} & \multicolumn{1}{c|}{\textbf{Label}} & \multicolumn{1}{c|}{\textbf{Modality}} & \multicolumn{1}{c|}{\textbf{Head-Pose}} & 
\multicolumn{1}{c|}{\textbf{Gaze}} &
\multicolumn{1}{c|}{\textbf{Env.}} & 
\multicolumn{1}{c|}{\textbf{Baseline}} & 
\multicolumn{1}{c|}{\textbf{\# Data}} & 
\multicolumn{1}{c}{\textbf{Year}} \\ \hline \hline
\href{https://www.cs.columbia.edu/CAVE/databases/columbia_gaze/}{CAVE}~\cite{smith2013gaze} & 56 & 3-D & \begin{tabular}[c]{@{}l@{}}Image\\ Dim.:$5184\times 3456$ \end{tabular} & $ 0$\textdegree, $\pm 30$\textdegree& $\pm 15$\textdegree, $\pm 10$\textdegree & In  & \begin{tabular}[c]{@{}l@{}}SVM\\ \textbf{Eval.:}Cross-val \end{tabular} &  \begin{tabular}[c]{@{}l@{}} \textbf{Total:}5880 \end{tabular}  &2013 \\ \hline 
\href{https://www.idiap.ch/en/dataset/eyediap}{EYEDIAP}~\cite{funes2014eyediap}  & 16  & 3-D &\begin{tabular}[c]{@{}l@{}} Image \\ Dim.: HD and VGA  \end{tabular}& $\pm 15$\textdegree, $30$\textdegree & $\pm 25$\textdegree, $ 20$\textdegree & In & \begin{tabular}[c]{@{}l@{}}Convex Hull \\ \textbf{Eval.:}  Hold out\end{tabular} & \begin{tabular}[c]{@{}l@{}} \textbf{Total:}237 min \end{tabular}&  2014\\ \hline
\href{https://www.ut-vision.org/datasets/}{UT MV}~\cite{sugano2014learning} & 50  & 3-D & \begin{tabular}[c]{@{}l@{}}Image \\Dim.:$1280\times 1024$ \end{tabular} & $\pm 36$\textdegree, $\pm 36$\textdegree& $\pm 50$\textdegree, $\pm 36$\textdegree & In & \begin{tabular}[c]{@{}l@{}}Random Reg.\\ Forests\\ \textbf{Eval.:}Hold out \end{tabular} & \begin{tabular}[c]{@{}l@{}} \textbf{Total:}64,000 \end{tabular}  & 2014\\ \hline 
OMEG~\cite{he2015omeg}& 50 & 3-D & \begin{tabular}[c]{@{}l@{}}Image\\Dim.: $1280 \times 1024$ \end{tabular} & $ 0$\textdegree, $\pm 30$\textdegree & \begin{tabular}[c]{@{}l@{}}$-38$\textdegree to $+36$\textdegree,\\ $-10$\textdegree to $+29$\textdegree \end{tabular} &In & \begin{tabular}[c]{@{}l@{}}SVR\\ \textbf{Eval.:}LOSO \end{tabular} & \begin{tabular}[c]{@{}l@{}} \textbf{Total:}44,827\end{tabular} &  2015\\ \hline
\href{https://www.mpi-inf.mpg.de/departments/computer-vision-and-machine-learning/research/gaze-based-human-computer-interaction/appearance-based-gaze-estimation-in-the-wild}{MPIIGaze}~\cite{zhang2015appearance}& 15 & 3-D & \begin{tabular}[c]{@{}l@{}}Image\\Dim.: $1280 \times 720$ \end{tabular} & $\pm 15$\textdegree, $30$\textdegree & $\pm 20$\textdegree, $\pm 20$\textdegree  &In & \begin{tabular}[c]{@{}l@{}}CNN variant~\cite{zhang2015appearance} \\ \textbf{Eval.:}LOSO \end{tabular} & \begin{tabular}[c]{@{}l@{}} \textbf{Total:}213,659 \end{tabular} &  2015\\ \hline
\href{http://gazefollow.csail.mit.edu/index.html}{GazeFollow}~\cite{nips15_recasens}& 130,339 & 3-D & \begin{tabular}[c]{@{}l@{}}Image\\Dim.: Variable \end{tabular} &  Variable & Variable &Both & \begin{tabular}[c]{@{}l@{}}CNN variant~\cite{nips15_recasens} \\ \textbf{Eval.:}Hold out \end{tabular} & \begin{tabular}[c]{@{}l@{}} \textbf{Total:}122,143 \end{tabular} &  2015\\ \hline
\href{https://www.cl.cam.ac.uk/research/rainbow/projects/syntheseyes/}{SynthesEye}\cite{wood2015rendering} & NA  & 3-D & \begin{tabular}[c]{@{}l@{}}Image \\Dim.:$120\times 80$ \end{tabular} & $\pm 50$\textdegree, $\pm 50$\textdegree& $\pm 50$\textdegree, $\pm 50$\textdegree & Syn & \begin{tabular}[c]{@{}l@{}}CNN~\cite{wood2015rendering}\\ \textbf{Eval.:}Hold out \end{tabular} & \begin{tabular}[c]{@{}l@{}} \textbf{Total:}11,400 \end{tabular}  & 2015\\ \hline
\href{https://gazecapture.csail.mit.edu/}{GazeCapture}~\cite{krafka2016eye}& 1450  & 2-D & \begin{tabular}[c]{@{}l@{}}Image\\ Dim.:$640\times 480$ \end{tabular} & $\pm 30$\textdegree, $40$\textdegree& $\pm 20$\textdegree, $\pm 20$\textdegree & Both & \begin{tabular}[c]{@{}l@{}}CNN~\cite{krafka2016eye}\\  \textbf{Eval.:} Hold out \end{tabular} & \begin{tabular}[c]{@{}l@{}} \textbf{Total:}2,445,504 \end{tabular} &  2016\\ \hline 
\href{https://www.cl.cam.ac.uk/research/rainbow/projects/unityeyes/}{UnityEyes}~\cite{wood2016learning}& NA  & 3-D & \begin{tabular}[c]{@{}l@{}}Image\\ Dim.:$400\times 300$ \end{tabular} & Variable & Variable & Syn & \begin{tabular}[c]{@{}l@{}}KNN \\  \textbf{Eval.:}NA \end{tabular} & \begin{tabular}[c]{@{}l@{}} \textbf{Total:} 1,000,000 \end{tabular} &  2016\\ \hline
\href{https://sh.rice.edu/cognitive-engagement/\%20tabletgaze/}{TabletGaze}~\cite{huang2015tabletgaze}& 51 & 2-D Sc. & \begin{tabular}[c]{@{}l@{}}Video \\ Dim.: $1280 \times 720$  \end{tabular} & $\pm 50$\textdegree, $\pm 50$\textdegree &$\pm 20$\textdegree, $\pm 20$\textdegree  & In & \begin{tabular}[c]{@{}l@{}}SVR\\ \textbf{Eval.:}Cross-val \end{tabular} & \begin{tabular}[c]{@{}l@{}} \textbf{Total:}816 Seq.\\ $\sim $ 300,000 img.\end{tabular} &2017\\ \hline %
\begin{tabular}[c]{@{}l@{}}\href{https://www.mpi-inf.mpg.de/departments/computer-vision-and-machine-learning/research/gaze-based-human-computer-interaction/its-written-all-over-your-face-full-face-appearance-based-gaze-estimation}{MPIIFaceGaze}\\~\cite{zhang2017s}\end{tabular}& 15 & 3-D & \begin{tabular}[c]{@{}l@{}}Image\\Dim.: $1280 \times 720$ \end{tabular} & $\pm 15$\textdegree, $30$\textdegree & $\pm 20$\textdegree, $\pm 20$\textdegree  &In & \begin{tabular}[c]{@{}l@{}}CNN variant~\cite{zhang2017s} \\ \textbf{Eval.:}LOSO \end{tabular} & \begin{tabular}[c]{@{}l@{}} \textbf{Total:}213,659 \end{tabular} &  2017\\ \hline
\href{https://www.mpi-inf.mpg.de/departments/computer-vision-and-machine-learning/research/gaze-based-human-computer-interaction/invisibleeye-mobile-eye-tracking-using-multiple-low-resolution-cameras-and-learning-based-gaze-estimation}{InvisibleEye}~\cite{tonsen2017invisibleeye}&  17 & 2-D Sc & \begin{tabular}[c]{@{}l@{}}Image \\ Dim.: $5 \times 5$ \end{tabular} & Unknown &\begin{tabular}[c]{@{}l@{}} $ 2560 \times 1600$\\pixel VF\end{tabular} & In & \begin{tabular}[c]{@{}l@{}}ANN~\cite{tonsen2017invisibleeye}\\ \textbf{Eval.:} Hold out\end{tabular} & \begin{tabular}[c]{@{}l@{}} \textbf{Total:}280,000\end{tabular} &2017\\ \hline
\href{https://github.com/Tobias-Fischer/rt_gene}{RT-GENE}~\cite{FischerECCV2018}&15  &3-D & \begin{tabular}[c]{@{}l@{}}Image \\Dim.:$1920\times 1080$ \end{tabular} &$\pm 40$\textdegree, $\pm 40$\textdegree & $\pm 40$\textdegree, $- 40$\textdegree &In & \begin{tabular}[c]{@{}l@{}}CNN~\cite{FischerECCV2018}\\ \textbf{Eval.:}Cross val \end{tabular} & \begin{tabular}[c]{@{}l@{}} \textbf{Total:}122,531\end{tabular}  & 2018\\ \hline
\href{http://gaze360.csail.mit.edu/}{Gaze 360}~\cite{gaze360_2019} & 238  & 3-D & \begin{tabular}[c]{@{}l@{}}Image \\ Dim.:$4096\times 3382$  \end{tabular} & $\pm 90$\textdegree, \small{u/k} & $\pm 140$\textdegree, $- 50$\textdegree &  \begin{tabular}[c]{@{}l@{}}Both  \end{tabular} & \begin{tabular}[c]{@{}l@{}}Pinball LSTM \\ \textbf{Eval.:} Hold out\end{tabular}& \begin{tabular}[c]{@{}l@{}} \textbf{Total:} 172,000 \end{tabular}& 2019 \\ \hline 
\href{https://github.com/Tobias-Fischer/rt_gene}{RT-BENE}\cite{cortacero2019rt} & 17  & EB & \begin{tabular}[c]{@{}l@{}}Image \\Dim.: $ 1920 \times 1080$ \end{tabular} & $\pm 40$\textdegree, $\pm 40$\textdegree  & $\pm 40$\textdegree, $- 40$\textdegree &  In & \begin{tabular}[c]{@{}l@{}}CNNs\\ \textbf{Eval.:} Cross val\end{tabular} & \begin{tabular}[c]{@{}l@{}} \textbf{Total:} 243,714\end{tabular} &2019\\ \hline
\href{https://sites.google.com/nvidia.com/nvgaze}{NV Gaze}~\cite{kim2019nvgaze} & 30  &\begin{tabular}[c]{@{}l@{}} 3-D,\\ Seg. \end{tabular}& \begin{tabular}[c]{@{}l@{}}Image (Synthetic) \\ Dim.:$1280 \times 960$, \\$640\times480 $ \end{tabular} & Unknown & \begin{tabular}[c]{@{}l@{}}$30$\textdegree $\times 40$\textdegree\\ VF \end{tabular}&  \begin{tabular}[c]{@{}l@{}}Both  \end{tabular} & \begin{tabular}[c]{@{}l@{}}CNN~\cite{laine2017production} \\ \textbf{Eval.:} Hold out\end{tabular}& \begin{tabular}[c]{@{}l@{}} \textbf{Total:} 2,500,000\\ \end{tabular}& 2019 \\ \hline %
\begin{tabular}[c]{@{}l@{}}\href{https://github.com/thorhu/Eyeblink-in-the-wild}{HUST-LEBW}\\~\cite{hu2019towards}\end{tabular} & 172  & EB & \begin{tabular}[c]{@{}l@{}}Video \\Dim.: $ 1280 \times 720$ \end{tabular} & Variable  & Variable &  Both & \begin{tabular}[c]{@{}l@{}}MS-LSTM\\ \textbf{Eval.:} Hold out\end{tabular} & \begin{tabular}[c]{@{}l@{}} \textbf{Total:} 673\end{tabular} &2019\\ \hline
\begin{tabular}[c]{@{}l@{}}\href{https://github.com/LifengFan/Human-Gaze-Communication}{VACATION}\\~\cite{fan2019understanding}\end{tabular} & 206,774  & GC & \begin{tabular}[c]{@{}l@{}}Video \\Dim.: $ 640 \times 360$ \end{tabular} & Variable  & Variable &  Both & \begin{tabular}[c]{@{}l@{}}GNN\\ \textbf{Eval.:} Hold out\end{tabular} & \begin{tabular}[c]{@{}l@{}} \textbf{Total:} 96,993\end{tabular} &2019\\ \hline
\begin{tabular}[c]{@{}l@{}}\href{https://research.facebook.com/openeds-challenge/}{OpenEDS-19}~\cite{garbin2019openeds} \\Track 1: Semantic\\ Segmentation \end{tabular}& 152  & Seg. &\begin{tabular}[c]{@{}l@{}}Image\\ Dim.: $640 \times 400$ \end{tabular} &Unknown &Unknown &  In &\begin{tabular}[c]{@{}l@{}}SegNet~\cite{badrinarayanan2017segnet}\\ \textbf{Eval.:} Hold out\end{tabular}  & \begin{tabular}[c]{@{}l@{}} \textbf{Total:}12,759\\(in \# SegSeq~\cite{garbin2019openeds}) \end{tabular}&2019\\ \hline 
\begin{tabular}[c]{@{}l@{}}\href{https://research.facebook.com/openeds-challenge/}{OpenEDS-19}~\cite{garbin2019openeds} \\Track 2: Synthetic \\ Eye Generation  \end{tabular} &  152 & Gen.  &\begin{tabular}[c]{@{}l@{}}Image \\ Dim.: $640 \times 400$\end{tabular} &Unknown & Unknown &  In & \begin{tabular}[c]{@{}l@{}}\textbf{Eval.:} Hold out\end{tabular} & \begin{tabular}[c]{@{}l@{}} \textbf{Total:} 252,690\end{tabular}&2019\\ \hline
\begin{tabular}[c]{@{}l@{}}\href{https://research.facebook.com/openeds-2020-challenge/}{OpenEDS-20}~\cite{palmero2020openeds2020} \\ Track 1: Gaze\\ Prediction \end{tabular} & 90  & 3-D   & \begin{tabular}[c]{@{}l@{}}Image\\ Dim.: $640 \times 400$ \end{tabular} & Unknown & $\pm 20$\textdegree, $\pm 20$\textdegree  &  In & \begin{tabular}[c]{@{}l@{}}Modified ResNet\\ \textbf{Eval.:} Hold out\end{tabular}  & \begin{tabular}[c]{@{}l@{}} \textbf{Total:} 8,960 Seq.,\\550,400 img.\end{tabular}&2020\\ \hline 
\begin{tabular}[c]{@{}l@{}}\href{https://research.facebook.com/openeds-2020-challenge/}{OpenEDS-20}~\cite{palmero2020openeds2020}\\ Track 2: Sparse \\ Temporal Semantic\\ Segmentation\end{tabular} & 90 & Seg.   & \begin{tabular}[c]{@{}l@{}}Image\\ Dim.: $640 \times 400$ \end{tabular} & Unknown & $\pm 20$\textdegree, $\pm 20$\textdegree &  In & \begin{tabular}[c]{@{}l@{}}\\ \textbf SegNet~\cite{badrinarayanan2017segnet}\\(Power \\Efficient version)\\{Eval.:} Hold out\end{tabular} & \begin{tabular}[c]{@{}l@{}} \textbf{Total:} 200 Seq.\\ 29,500 img.\end{tabular} &2020\\ \hline 
\href{https://github.com/BiDAlab/mEBAL}{mEBAL}\cite{daza2020mebal} & 38  & EB & \begin{tabular}[c]{@{}l@{}}Image \\Dim.: $ 1280 \times 720$ \end{tabular} & Variable  & Variable &  In & \begin{tabular}[c]{@{}l@{}}VGG-16 Varient\\ \textbf{Eval.:} Hold out\end{tabular} & \begin{tabular}[c]{@{}l@{}} \textbf{Total:} 756,000\end{tabular} &2020\\ \hline
\href{https://ait.ethz.ch/projects/2020/ETH-XGaze/}{ETH-XGaze}~\cite{zhang2020eth} & 110  & 3-D & \begin{tabular}[c]{@{}l@{}}Image \\Dim.: $6000\times 4000$ \end{tabular} & $\pm 80$\textdegree, $\pm 80$\textdegree  & $\pm 120$\textdegree, $\pm 70$\textdegree&  In & \begin{tabular}[c]{@{}l@{}}ResNet-50\\ \textbf{Eval.:} Hold out\end{tabular} & \begin{tabular}[c]{@{}l@{}} \textbf{Total:} 1,083,492\end{tabular} &2020\\ \hline 
\href{https://ait.ethz.ch/projects/2020/EVE/}{EVE}~\cite{Park2020ECCV} & 54  & 3-D & \begin{tabular}[c]{@{}l@{}}Image \\Dim.: $6000\times 4000$\\ \end{tabular} & $\pm 80$\textdegree, $\pm 80$\textdegree  & $\pm 80$\textdegree, $\pm 80$\textdegree&  In & \begin{tabular}[c]{@{}l@{}}ResNet-18\\ \textbf{Eval.:} Hold out\end{tabular} & \begin{tabular}[c]{@{}l@{}} \textbf{Total:} 12,308,334\end{tabular} &2020\\ \hline 
\href{http://www.cis.rit.edu/~rsk3900/gaze-in-wild/}{GW}~\cite{kothari2020gaze} & 19  & GE & \begin{tabular}[c]{@{}l@{}}Image \\Dim.: $1920\times 1080$\\ \end{tabular} & Variable  & Variable &  In & \begin{tabular}[c]{@{}l@{}}RNN\\ \textbf{Eval.:} Hold out\end{tabular} & \begin{tabular}[c]{@{}l@{}} \textbf{Total:} $\sim$ 5,800,000\end{tabular} &2020\\ \hline
\href{https://github.com/AVAuco/ucolaeodb}{LAEO}~\cite{Kothari_2021_CVPR} & 485  & 3-D & \begin{tabular}[c]{@{}l@{}}Image \\Dim.: Variable\\ \end{tabular} & Variable  & Variable &  Both & \begin{tabular}[c]{@{}l@{}}ResNet-18+LSTM\\ \textbf{Eval.:} Hold out\end{tabular} & \begin{tabular}[c]{@{}l@{}} \textbf{Total:} 800,000\end{tabular} &2021\\ \hline
\href{https://github.com/upeee/GOO-GAZE2021}{GOO}~\cite{tomas2021goo} & 100  & 3-D & \begin{tabular}[c]{@{}l@{}}Image \\Dim.: Variable\\ \end{tabular} & Variable  & Variable &  Both & \begin{tabular}[c]{@{}l@{}}ResNet-50\\ \textbf{Eval.:} Hold out\end{tabular} & \begin{tabular}[c]{@{}l@{}} \textbf{Total:} 201,552\end{tabular} &2021\\ \hline
\href{https://research.facebook.com/publications/openneeds-a-dataset-of-gaze-head-hand-and-scene-signals-during-exploration-in-open-ended-vr-environments/}{OpenNEEDS}~\cite{emery2021openneeds} & 44  & 3-D & \begin{tabular}[c]{@{}l@{}}Image \\Dim.: $128\times 71$\\ \end{tabular} & Variable  & Variable &  VR & \begin{tabular}[c]{@{}l@{}}GBRT\\ \textbf{Eval.:} Hold out\end{tabular} & \begin{tabular}[c]{@{}l@{}} \textbf{Total:} 2,086,507\end{tabular} &2021\\ 
\bottomrule[0.4mm]
\end{tabular}}
\vspace{-5mm}
\end{table*}

\begin{table*}[!htbp]
\caption{\textbf{Cross-Dataset Study.} Cross dataset generalization study on different gaze estimation datasets in terms of angular error (in \textdegree).}
\label{tab:cross_dataset} 
\centering
\begin{tabular}{l|c|c}
\toprule[0.4mm]
\rowcolor{mygray}
\multicolumn{1}{l|}{\textbf{Model}} & \multicolumn{1}{c|}{\textbf{\begin{tabular}[c]{@{}l@{}}Test$\rightarrow$\\ Train\\$\downarrow$\end{tabular}}} & \multicolumn{1}{c}{\begin{tabular}[c]{@{}l@{}}\textbf{Datasets}\end{tabular}} \\ \hline \hline
Pinball-LSTM~\cite{gaze360_2019} & \begin{tabular}[c]{@{}c@{}} \\ CAVE\\MPIIFace \\RT-GENE \\Gaze360 \end{tabular}& \begin{tabular}[c]{c c c c} CAVE & MPIIFace & RT-GENE & Gaze360\\ \hline
--& 12.3\textdegree &  32.8\textdegree & 57.9\textdegree \\
12.4\textdegree & -- &  26.5\textdegree & 57.8\textdegree \\
24.2\textdegree & 18.9 &  -- & 56.6\textdegree \\
9.0\textdegree & 12.1 &  13.4\textdegree & --\\
\end{tabular} \\ \hline
ETH-X Gaze~\cite{zhang2020eth} & \begin{tabular}[c]{@{}c@{}} \\ \\  MPIIGaze\\EYEDIAP \\Gaze-Capture\\RT-GENE \\Gaze360\\ETHXGaze \end{tabular}& \begin{tabular}[c]{c c c c c c} MPIIGaze & EYEDIAP & Gaze-Capture & RT-GENE & Gaze360 & ETH-X Gaze\\ \hline
--& 17.9\textdegree &  6.3\textdegree & 14.9\textdegree & 31.7\textdegree & 34.9\textdegree\\
16.9\textdegree & -- &  14.2\textdegree & 15.6\textdegree & 33.7\textdegree& 41.7\textdegree \\
4.5\textdegree & 13.7\textdegree &  -- & 14.7\textdegree & 30.2\textdegree& 29.4\textdegree \\
12.0\textdegree & 21.2\textdegree &  13.2\textdegree& -- & 34.7\textdegree& 42.6\textdegree \\
10.3\textdegree & 11.3\textdegree &  12.9\textdegree & 26.6\textdegree & --& 17.0\textdegree \\
7.5\textdegree & 11.0\textdegree &  10.5\textdegree& 31.2\textdegree & 27.3\textdegree& --
\end{tabular} \\ 
\bottomrule[0.4mm]
\end{tabular}
\end{table*}

\noindent \textbf{Data Generation/Gaze Redirection.}
Since gaze data collection and annotation is an expensive and time-consuming process, the research community moves towards a data generation process for benchmarking with a large variation in data attributes. Prior works in this domain generate both synthetic and real images. The methods are based on Generative Adversarial Networks (GANs). To capture the possible rotational variation in images, gaze redirection techniques~\cite{ganin2016deepwarp,he2019photo,wood2018gazedirector,yu2019improving,kaur2021subject} are quite popular. An early work on gaze manipulation~\cite{wolf2010eye} uses pre-recording of several potential eye replacements during test time. Further, Kononenko et al.~\cite{kononenko2015learning} propose wrapping based gaze redirection using supervised learning, which learns the gaze redirection via a flow field to move eye pupil and relevant pixels from the input image to the output image. The gaze re-direction methods may struggle with extrapolation since it depends on the training samples and training methods. Moreover, these works suffer from low-quality generation and low redirection precision. To overcome this, Chen et al.~\cite{chen2020mggr} propose a MultiModal-Guided Gaze Redirection (MGGR) framework which uses gaze-map images and target angles to adjust a given eye appearance via learning. The other approaches are mainly based on random forest~\cite{kononenko2015learning} and style transfer~\cite{sela2017gazegan}. Random forest is used to decide the possible gaze direction and in style transfer, the appearance based feature is mainly encoded. Sela et al.~\cite{sela2017gazegan} propose a GAN based framework to generate a large dataset of high-resolution eye images having diversity in subjects, head pose, camera settings and realism. However, the GAN based methods lack in their capability to preserve content (i.e. eye shape) for benchmarking. Buhler et al.~\cite{buhler2019content} synthesize person-specific eye images with a given semantic segmentation mask by preserving the style and content of the reference images. In summary, we can say that although a lot of effort has been made to generate realistic eye images, but due to several limitations (perfect gaze direction, image quality), these images are not used for benchmarking.

\subsection{Evaluation Strategy}

In this section, we describe the most widely used gaze metrics in the gaze analysis domain. 

\noindent \textbf{Gaze Estimation.} The most common practice to measure the gaze estimation accuracy/error is in terms of angular error (in \textdegree)~\cite{park2018deep,zhang2020eth,park2019few,Park2020ECCV} and gaze location (in pixels or cm/mm(s))~\cite{huang2015tabletgaze,Park2020ECCV}. The angular error is measured between the actual gaze direction ($\mathbf{g}\in \mathbb{R}^3$) and predicted gaze direction ($\hat{\mathbf{g}}\in \mathbb{R}^3$) defined as $\frac{\mathbf{g}.\hat{\mathbf{g}}}{||\mathbf{g}||.||\hat{\mathbf{g}}||}$. On the other hand, Euclidean distance is measured between the original and predicted point of gaze (PoG).

\noindent \textbf{Gaze Redirection.} The gaze redirection evaluation is performed in both quantitative and qualitative manner~\cite{zheng2020self,chen2020mggr,chen2021coarse}. The quantitative analysis is done in terms of angular gaze redirection error estimated between the predicted values and their intended target values. As in this task, the moment of the eye pupil is pre-defined, thus, this angular error weakly quantifies how perfectly the eye redirection occurs, although the method for measuring the angle has some inherent noise. 
For qualitative analysis, the Learned Perceptual Image Patch Similarity (LPIPS) metric is used which measures the paired image similarity in the gaze redirection task.

\noindent \textbf{Eye Segmentation.} Commonly used evaluation metric for eye segmentation methods, is average of the mean Intersection over
Union (mIoU). Although, for the recent OpenEDS challenge~\cite{garbin2019openeds}, the mIoU metric is calculated for all classes and model size (S) is calculated as a function of a number of trainable parameters in megabytes (MB).

\subsection{Cross Dataset Analysis.}
Datasets play an important role in defining the research progress made in gaze analysis. Apart from serving as a source for training models, it helps to quantify the performance measure. In the gaze analysis domain, the aim of dataset curation is to capture the real-world scenario setting as close as possible. Thus, it is necessary to evaluate the robustness and generalizability of the models across different data acquisition setups for better adaptation.

\noindent On this front, we explore two aspects: First of all, we explore the cross-dataset generalizability of gaze estimation methods based on two SOTA models i.e. Pinball-LSTM~\cite{gaze360_2019} and ETH-X-Gaze~\cite{zhang2020eth}. For this purpose, the training is performed on one dataset while the testing is conducted on the other dataset. (Refer Table~\ref{tab:cross_dataset}). Further, we explore the SOTA method's performance on different datasets to show the robustness of the model (Refer Table~\ref{tab:where_we_stand}). Angular error (in \textdegree) is used as an evaluation metric. Further to generalize across datasets, we calculate the mean angular error across datasets. Below are some of the important observations inferred from our experimental results.


\noindent \textbf{Data Collection Settings.} Dataset collection setup plays an important role in generalizability and method's robustness. As the CAVE dataset is collected in a constrained setup and it has high-resolution images, the models trained on this data fail to adapt well to the data with low resolution and synthetic images. Thus, the pinball-LSTM trained on CAVE data has a high error in Gaze360 ($\sim$ 57.9\textdegree) and RT-GENE ($\sim$ 32.8\textdegree) datasets. A similar pattern is observed in the case of the MPII dataset as well. Model trained on this dataset gives high error in adapting RT-GENE ($\sim$ 26.5\textdegree) and Gaze360 ($\sim$ 57.8\textdegree).

\noindent \textbf{Cross Dataset Generalization.} By observing the cross dataset generalization performance, we can determine how diverse the training dataset is from a generalization perspective. From Table~\ref{tab:cross_dataset}, we observe that Gaze360, Gaze-Capture, and ETH-X Gaze datasets are the most challenging datasets. Training models on these two datasets would be a good choice as it has better generalization performance across different datasets. In contrast, RT-GENE and Gaze-capture contain significant biases and training models on them will lead to poor cross-dataset generalization performance.

\noindent \textbf{Robust Modelling.} In order to study the robustness of a model trained on any dataset, we recommend evaluating the model on Gaze360, Gaze-Capture, and ETH-X Gaze datasets. These datasets exhibit multiple variations in terms of background environment, eye visibility, occlusion, low-resolution images and could serve as important indicators for real-world adaptation. We also recommend training the models on all benchmark datasets together and expect a better generalization than training on individual datasets in novel or in-the-wild settings.


\begin{table*}[!htbp]
\caption{\textbf{Where we stand now.} Chronological comparison of the performance of different models for the gaze-related tasks on related benchmark datasets.}
\label{tab:where_we_stand} 
\centering
\begin{tabular}{l|c|c|c}
\toprule[0.4mm]
\rowcolor{mygray}
\multicolumn{1}{l|}{\textbf{Task}} &
\multicolumn{1}{c|}{\textbf{Methods}} & \multicolumn{1}{c|}{\textbf{Datasets}} & \multicolumn{1}{c}{\textbf{Year}}\\ \hline \hline
\begin{tabular}[l]{@{}l@{}} Gaze\\Estimation \end{tabular}& \multicolumn{1}{l|}{\begin{tabular}[l]{@{}l@{}} \\GazeNet~\cite{zhang2015appearance} \\Dilated-Conv.~\cite{chen2018appearance}\\ Landmark based~\cite{park2018learning}\\ RT-GENE~\cite{FischerECCV2018} \\ Pinball-LSTM~\cite{gaze360_2019} \\ CA-Net~\cite{cheng2020coarse}\\ \href{https://github.com/yihuacheng/GazeTR}{GazeTR-Hybrid}~\cite{cheng2021gaze}\end{tabular}} & \begin{tabular}[c]{c c c c c c c} CAVE & MPIIGaze & EYEDIAP & UT MV & MPIIFace & Gaze360 & ETH-X Gaze\\ \hline
-- & 5.70\textdegree & 7.13\textdegree& 6.44\textdegree & 5.76\textdegree & -- & -- \\
--& 4.39\textdegree & 6.57\textdegree & -- & 4.42\textdegree &13.73\textdegree & --\\
8.7\textdegree & 8.3\textdegree & 26.6\textdegree& -- & --& -- & -- \\
-- & 4.61\textdegree & 6.30\textdegree& -- & 4.66\textdegree & 12.26 & -- \\
9.0\textdegree & 12.1\textdegree & 5.58\textdegree& -- & 12.1\textdegree & 11.04\textdegree & 4.46\textdegree \\
-- & 4.27\textdegree & 5.63\textdegree& -- & 4.27\textdegree & 11.20\textdegree & --\\
\end{tabular} & \begin{tabular}[c]{@{}c@{}} \\ 2015 \\ 2018 \\ 2018\\ 2019 \\2020\\2021\end{tabular}\\\hline
\multicolumn{4}{l}{\begin{tabular}[c]{@{}c@{}} CAVE \hspace{35mm} MPIIGaze \hspace{35mm} Eth-X-Gaze \hspace{35mm} Gaze360  \\ \frame{\includegraphics[width=45mm, height=25mm]{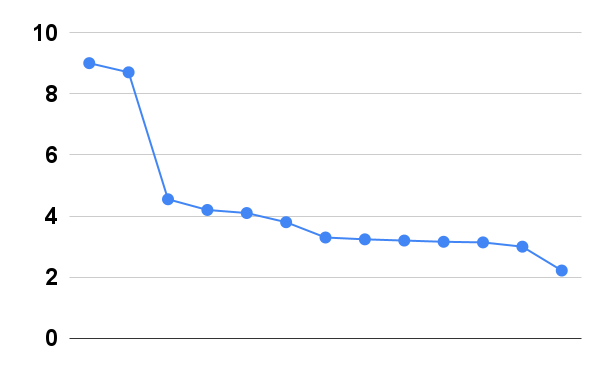}}
\frame{\includegraphics[width=45mm, height=25mm]{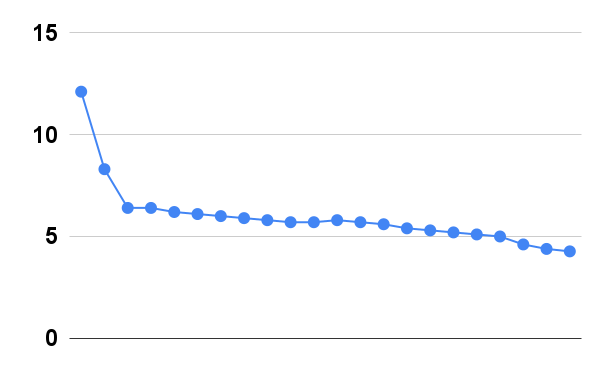}}
\frame{\includegraphics[width=45mm, height=25mm]{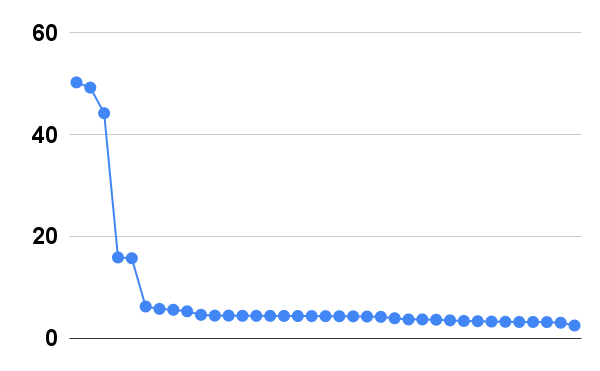}}
\frame{\includegraphics[width=45mm, height=25mm]{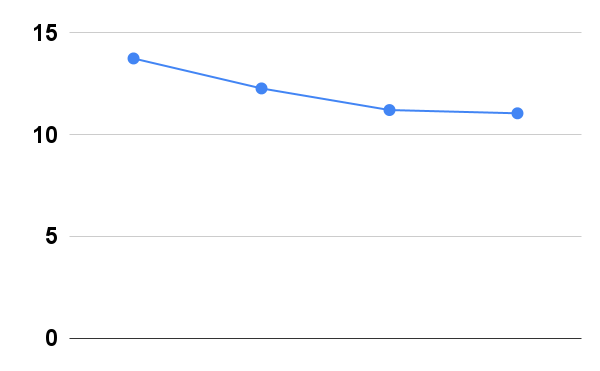}}
\end{tabular} }\\
\multicolumn{4}{c}{In the diagrams, the y-axis represents the angular error (in\ \textdegree) and the x-axis represents the timeline.}

\\ \hline \hline
\rowcolor{mygray}
\multicolumn{1}{l|}{\textbf{Task}} &
\multicolumn{1}{c|}{\textbf{Datasets}} & \multicolumn{1}{c|}{\textbf{Methods (Eval.AI)}} & \multicolumn{1}{c}{\textbf{Year}}\\ \hline \hline
\begin{tabular}[l]{@{}l@{}} Trajectory\\Estimation \end{tabular}& \multicolumn{1}{l|}{\begin{tabular}[l]{@{}l@{}} Team-name $\rightarrow$\\OpenEDS2020~\cite{palmero2020openeds2020}\\
Team-name $\rightarrow$ \\OpenNEEDS~\cite{emery2021openneeds}\end{tabular}} & \begin{tabular}[c]{c c c c c c } Random\_B & caixin & EyMazing & fgp200709d & vipl\_gaze & Baseline\\ \hline
\href{https://repushko.com/all/openeds2020/}{3.078\textdegree} & 3.248\textdegree & 3.313\textdegree & 3.347\textdegree & 3.386\textdegree &  5.368\textdegree \\
XiaodongWang & Hebut\_Lyx & tetelias & TCS\_Research & Baseline &AnotherShot\\ \hline
1.68\textdegree & 1.75\textdegree & 1.99\textdegree & 2.05\textdegree & 7.18\textdegree & 7.94\textdegree \end{tabular} & \begin{tabular}[c]{@{}c@{}} \\ \href{https://eval.ai/web/challenges/challenge-page/605/leaderboard/1682}{2020}\\ \\ \href{https://eval.ai/web/challenges/challenge-page/895/leaderboard/2361}{2021}\end{tabular} 

\\\hline \hline
\rowcolor{mygray}
\multicolumn{1}{l|}{\textbf{Task}} &
\multicolumn{1}{c|}{\textbf{Dataset}} & \multicolumn{1}{c|}{\textbf{Methods}} & \multicolumn{1}{c}{\textbf{Year}}\\ \hline \hline
\begin{tabular}[l]{@{}l@{}} Gaze\\Zone \end{tabular}& \multicolumn{1}{l|}{\begin{tabular}[l]{@{}l@{}}
\href{http://users.cecs.anu.edu.au/~Tom.Gedeon/pdfs/Emotiw\%202020\%20Driver\%20gaze\%20group\%20emotion\%20student\%20engagement\%20and\%20physiological\%20signal\%20based\%20challenges.pdf}{EmotiW2020} $\rightarrow$\\DGW~\cite{ghosh2020speak2label} \end{tabular}} & \begin{tabular}[c]{c c c c c c c} DD\_Vision & SituAlgorithm & Overfit & DeepBlueAI & UDECE & X-AWARE & Baseline\\ \hline
82.52\% & 81.51\% & 78.87\% & 75.88\% & 74.57\% & 71.62\% & 60.98\% \\
\end{tabular} & \begin{tabular}[c]{@{}c@{}} \\2020\end{tabular} 

\\\hline \hline
\rowcolor{mygray}
\multicolumn{1}{l|}{\textbf{Task}} &
\multicolumn{1}{c|}{\textbf{Datasets}} & \multicolumn{1}{c|}{\textbf{Methods}} & \multicolumn{1}{c}{\textbf{Year}}\\ \hline \hline
\begin{tabular}[l]{@{}l@{}} Visual\\Attention \end{tabular}& \multicolumn{1}{l|}{\begin{tabular}[l]{@{}l@{}} \\GazeFollowing~\cite{NIPS2015_ec895663} (AUC $\uparrow$)\\
\\GazeCommunication~\cite{fan2019understanding}\\
\textit{1. Atomic-Level} (Top-1\%)\\
\textit{2. Event-Level} (Top-1\%)\\ \\
VisualSearch~\cite{zelinsky2019benchmarking}\\
\textit{1. Microwave} (MultiMatch)\\
\textit{2. Clock} (MultiMatch)\\
\\ SharedAttention~\cite{chong2020detecting}\end{tabular}} & \begin{tabular}[c]{c c c c c c} 
Human & \cite{chong2020detecting} & \cite{chong2018connecting} &  \cite{NIPS2015_ec895663} & Center & Random\\ \hline
0.924& 0.921 & 0.896 &0.878  & 0.633 & 0.504\\ 

ST-GNN & CNN+LSTM & CNN+SVM &  CNN+RF & CNN & Chance\\ \hline
 55.02\% & 24.65\% &  36.23\% & 37.68\% & 23.05\% & 16.44\% \\
55.90\%& -- & -- & -- &-- & 22.70\%\\ 

\begin{tabular}[c]{@{}c@{}}Behavioural\\Agreement\end{tabular} & CNN & \begin{tabular}[c]{@{}c@{}}RNN \end{tabular}&  LSTM & GRU & Scanpath \\ \hline
 0.714 &  0.621 & 0.677 & 0.684 & 0.664 & Direction\\ 
 0.701 &  0.633 & 0.673 & 0.669 & 0.659 & Direction\\ 

\begin{tabular}[c]{@{}c@{}}ST-CNN\\+LSTM\end{tabular} & ST-GNN & \begin{tabular}[c]{@{}c@{}}Gaze+Saliency\\+LSTM \end{tabular}&  Gaze+Saliency & GazeFollow &  Random\\ \hline
83.3\%& 71.4\% & 66.2\% & 59.4\% & 58.7\% & 22.70\%\\ 
\end{tabular} & \begin{tabular}[c]{@{}c@{}} \\ 2015\\ \\ \\2019\\ \\ \\ \\2019\\ \\ \\2020\\\end{tabular} 
\\
\bottomrule[0.4mm]
\end{tabular}
\end{table*}

\subsection{Where We Stand Now?}
In Table~\ref{tab:where_we_stand}, we analyze dataset specific improvements made by different methods over the past few years. In the following, we discuss some of the important observations from Table~\ref{tab:where_we_stand}.

\noindent \textbf{Gaze Estimation on Constrained Setup.} Most popular \textit{2D-3D gaze estimation} datasets~\cite{zhang2017mpiigaze,zhang2020eth,smith2013gaze} are collected in constrained scenarios where there is a certain distance between the user and the visual screen. Moreover, as the nodal point of the human eye has subject-specific offset (which varies around 2-3\textdegree), it is difficult to reduce the angular error beyond a certain limit using visible regions of the eyes. On this front, the performance of some of the gaze estimation methods~\cite{zhang2015appearance,zhang2017s,zhang2020eth} seems to have plateaued on constrained datasets such as CAVE, MPII, and Eth-X-Gaze (Refer to Figures in Table~\ref{tab:where_we_stand}). 

\noindent \textbf{Gaze Estimation in Unconstrained Setup.} Gaze estimation in unconstrained environments still remains largely unresolved mainly due to the unavailability of large-scale annotated data. Gaze360~\cite{gaze360_2019} and GazeCapture~\cite{krafka2016eye} are two popular public-domain datasets available for this purpose. Especially in the Gaze360 dataset, in many cases, the eyes are not visible which makes it more challenging to track where the person is looking. It is quite difficult to estimate gaze in a naturalistic environment, more exploration along this line is highly desirable.

\noindent \textbf{Gaze Estimation with Limited Supervision.} Gaze Estimation with limited supervision is a promising research direction. As manual annotation of gaze data is an error-prone process, there is a high possibility of noise in labeling. To this end, proposed approaches are mainly based on `learning-by-synthesis'~\cite{sugano2014learning}, hierarchical generative models~\cite{wang2018hierarchical}, conditional random field~\cite{benfold2011unsupervised}, unsupervised gaze target discovery~\cite{zhang2017everyday}, few-shot learning~\cite{park2019few,yu2019improving}, pseudo-labelling~\cite{ghosh2022mtgls} and self/unsupervised~\cite{yu2019unsupervised,dubey2019unsupervised}. While domain specific knowledge has been utilized for these approaches, developing robust methods from limited amount of annotated data with enhanced generalization across different real-life scenarios still largely remains unresolved.

\noindent \textbf{Visual Attention Estimation.} Eye visibility plays an important role in estimating the gaze direction of a person. To this end, visual attention estimation mainly focuses on where the person is looking irrespective of eye visibility. To facilitate research along this direction, GazeFollow~\cite{NIPS2015_ec895663} and VideoAttentionTarget~\cite{chong2020detecting} datasets have been proposed. Some important research directions to explore include scene saliency, visual search, and human scan path~\cite{yang2020predicting,chen2021predicting}. 

\noindent \textbf{Gaze Trajectory Modelling.} Gaze Trajectory modeling and estimation is another line of research that requires further research attention~\cite{palmero2020openeds2020}. Natural gaze dynamics consist of a continuous sequence of gaze events such as \textit{fixations, saccades, pursuit, vestibulo-ocular reflex, optokinetic reflex, vergence,} and \textit{blinks}~\cite{rayner1995eye}. These aforementioned dynamics can be influenced by saliency, task-relevant information, and environmental factors. Natural eye movements of humans span an elliptical region with a horizontally oriented axis greater than $\sim 100$\textdegree\ and a vertically oriented axis spanning $\sim 70$\textdegree. Due to the lack of labeled temporal gaze trajectory data, there are only a few studies~\cite{palmero2020openeds2020,palmero2020benefits} that focus on tasks related to the gaze trajectory.


\section{Applications} \label{sec:application}
\subsection{Gaze in Augmented Reality, Virtual Reality and 360\textdegree\ Video Streaming}
We are witnessing great progress in the adaptation of VR, AR and 360\textdegree\ Video Streaming technology. Eye-tracking has the potential to bring revolution in the AR/VR and 360\textdegree\ video streaming for immersive video application space since it can enhance the device's awareness by learning about users' attention at any given point in time. Consequently, user's focus based optimization reduces power consumption by these devices~\cite{10.1145/3308755.3308765,patney2016towards,palmero2020openeds2020,park2021mosaic}. In this section, we will cover the importance of eye-tracking technology and how it enables better user experience in AR/VR and 360\textdegree\ Video Streaming devices including eye inter pupillary distance for estimating image perception quality, person identification or state  estimation by their eye gaze pattern, improve interactions, etc. 

Foveated Rendering (a.k.a gaze-contingent eye tracking) is a process designed to show the user only a portion of what they are looking at in full detail~\cite{patney2016towards,garbin2019openeds,palmero2020openeds2020}. The focus region follows the user's visual field. Graphics displayed with foveated rendering better matches the way we see objects. Usually, the user watches the AR/VR environment or 360\textdegree\ video using head mounted display devices. The existing platforms stream the full 360\textdegree\ scene while the user can view only a small part of the scene which spans about 90\textdegree\ - 120\textdegree\ horizontally, 90\textdegree\ vertically. Quantitatively, it is less than 20\% of the whole scene. Thus, a significant amount of power and network bandwidth is wasted for the display which is never utilized in viewing. In ideal condition, the display will be only in the user's visual field while blurring the periphery. Following are the three important benefits of the user's visual field based rendering process: \textit{1. Improved Image Quality:} It can enable 4k displays on the current generation graphics processing units (GPUs) without degradation in performance. \textit{2. Lower cost:} Similarly, the end-users can run AR/VR and 360\textdegree\ Video Streaming based applications on low-cost hardware without compromising the performance. \textit{3. Increased Frame Rate per Second (FPS):} The end-user can run at a higher frame rate using the same graphical settings. There are two types of foveated rendering: \textit{dynamic foveated rendering} and \textit{static foveated rendering}. Dynamic foveated rendering follows the user’s gaze trajectory using eye-tracking and renders a sharp image in the required region, but this eye tracking is challenging in many scenarios. On the other hand, static foveated rendering considers a fixed area of the highest resolution at the center of the viewer’s device irrespective of the user’s gaze. It depends on the user’s head movements, thus, facing a challenge in eye-head interplay as the image quality is drastically reduced if the user looks away from the center of the field of view. 
The main key aspects of accurate \textit{eye position/visual attention} estimation ahead of time is to enhance user experience via providing high image quality in the subject's visual focus area. It requires person-specific calibration as nodal point of human have subject specific offset. Thus, generalizing it across user poses a challenge in the gaze analysis community to address~\cite{10.1145/3313831.3376260}. On the other hand, user's viewpoint prediction ahead of time could face a lot of challenges as human eye movement is ballistic in nature. The visual attention of the user can therefore change abruptly based on the content in the screen. Thus, the prediction algorithm needs to take care of imperfect prediction as well and it needs to integrate with bit rate control. This process will enable user-specific recommendation and other facilities to enhance user experience~\cite{10.1145/3277644.3277771}.

\subsection{Driver Engagement}
With the progress in autonomous and smart cars, the requirement for automatic driver monitoring has been observed and researchers have been working on this problem for a few years now~\cite{vasli2016driver,tawari2014driver,ghosh2020speak2label,jha2020multimodal}. In the literature, the problem is treated as a gaze zone estimation problem. A summary of the gaze estimation benchmarks is shown in Table \ref{tab:driver_gaze}. The proposed methods can be classified into two categories:

\noindent \textbf{Sensor Based Tracking.} These mainly utilize dedicated sensors integrated hardware devices for monitoring the driver's gaze in real-time. These devices require accurate pre-calibration and additionally these devices are expensive. Few examples of these sensors are Infrared (IR) camera~\cite{johns2007monitoring}, head-mounted devices~\cite{jha2018probabilistic,jha2017challenges} and other systems~\cite{feng2013low,zhang2017exploring}. 


\begin{table}[t]
\caption{Comparison of driver gaze estimation datasets with respect to number of subjects (\# Sub), number of zones (\# Zones), illumination conditions and labelling procedure.} 
\label{tab:driver_gaze} 
\centering
\begin{tabular}{l|c|c|c|c}
\toprule[0.4mm]
\rowcolor{mygray}
\multicolumn{1}{c|}{\textbf{References}} & \multicolumn{1}{c|}{\textbf{\# Sub}} & \multicolumn{1}{c|}{\textbf{\# Zones}} & \multicolumn{1}{c|}{\textbf{Illumination}} & \multicolumn{1}{c}{\textbf{Labelling}}  \\ \hline \hline

Choi et al.~\cite{choi2016real}   & 4                                                              & 8                                                               & \begin{tabular}[c]{@{}c@{}}Bright \&\\ Dim\end{tabular}        & 3-D Gyro.                                                      \\ \hline
\begin{tabular}[c]{@{}l@{}}Lee et al. ~\cite{lee2011real} \end{tabular} & 12                                                             & 18                                                               & Day                                                            & Manual                                                      \\ \hline
\begin{tabular}[c]{@{}l@{}}Fridman et al. ~\cite{fridman2015driver} \end{tabular} & 50                                                             & 6                                                               & Day                                                            & Manual                                                      \\ \hline
Tawari et al.  ~\cite{tawari2014driver}                                           & 6                                                              & 8                                                               & Day                                                            & Manual                                                      \\ \hline
Vora et al. ~\cite{vora2018driver}                                              & 10                                                             & 7                                                               & \begin{tabular}[c]{@{}c@{}}Diff. \\ day times\end{tabular} & Manual                                                      \\ \hline
Jha et al.~\cite{jha2018probabilistic}                                                 & 16                                                             & 18                                                              & Day                                                            & \begin{tabular}[c]{@{}c@{}}Head-\\ band\end{tabular} \\ \hline
Wang et al.~\cite{wang2019continuous}                                                &                                                           3     & 9                                                               & Day                                                            & \begin{tabular}[c]{@{}c@{}}Motion\\ Sensor\end{tabular}                                                  \\ \hline
\begin{tabular}[l]{@{}l@{}} DGW\cite{ghosh2020speak2label} \end{tabular}                                                     & 338                                                            & 9                                                               & \begin{tabular}[c]{@{}c@{}}Diff. day times\end{tabular}          & Automatic                                                     \\ \hline
\begin{tabular}[l]{@{}l@{}} MGM\cite{jha2020multimodal} \end{tabular}                                                     & 60                                                            & 21                                                               & \begin{tabular}[c]{@{}c@{}}Diff. day times \end{tabular}          & \begin{tabular}[c]{@{}c@{}}Multiple\\ Sensors \end{tabular}                                                    \\ 
\bottomrule[0.4mm]
\end{tabular}
\vspace{-5mm}
\end{table}

\noindent \textbf{Image processing and vision based methods.} These are mainly focused on two types of methods: head-pose based only~\cite{lee2011real,tawari2014robust,mukherjee2015deep,wang2019continuous} and both head-pose and eye-gaze based~\cite{vasli2016driver,tawari2014driver,tawari2014robust,fridman2015driver,fridman2016owl,choi2016real}. Driver's head pose provides partial information regarding his/her gaze direction as there may be an interplay between eyeball movement and head pose~\cite{fridman2016owl}. Hence, methods relying on head pose information may fail to disambiguate between the eye movement with fixed head-pose. Thus, the methods relying on both head pose and gaze based prediction are more robust.

\subsection{Gaze in Healthcare and Wellbeing}
Gaze is widely used in healthcare domain to enhance the diagnosis performance. Generally, eye movement patterns is widely used as behavioral bio-markers of various mental health problems including depression~\cite{alghowinem2013eye}, post traumatic stress disorder~\cite{milanak2018ptsd} and Parkinson’s disease~\cite{harezlak2018application}. Similarly, individuals diagnosed with Autism Spectral Disorder display gaze avoidance in social scenes~\cite{harezlak2018application}. Even intoxication including alcohol consumption and/or other drugs usage reflects on eye and gaze properties, especially, decreased accuracy and speed of saccades, changes in pupil size, and an impaired ability to fixate on moving objects. A recent survey~\cite{harezlak2018application} discusses the potential applications in healthcare including concussion~\cite{kempinski2016system}, multiple sclerosis~\cite{avital2015method}.

\begin{figure}[t]
    \centering
    \includegraphics[width= \linewidth]{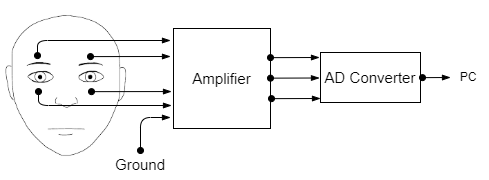}
    \caption{Electro-oculogram (EOG) based gaze estimation method~\cite{sakurai2017gaze}. This prototype opens the possibility of communication for severely disabled people. Refer Sec.~\ref{sec:physiological_signals} for more details. }
    \label{fig:gaze_sensor}
\end{figure}

\noindent \textbf{Physiological Signals.}\label{sec:physiological_signals}
A gaze estimation system could be one of the communication methods for severely disabled people who cannot perform any type of gestures and speech. Sakurai et al.~\cite{sakurai2016study} developed an eye-tracking method using a compact and light electrooculogram (EOG) signal. Further, this prototype is improved via the usage of the EOG component which strongly correlated with the change of eye movements~\cite{sakurai2017gaze} (Refer Fig.~\ref{fig:gaze_sensor}). The setup can detect object scanning only by eye and face muscle movements. The experimental results open the possibility of eye-tracking via EOG signals and a Kinect sensor. Research along this direction can be extremely useful for disabled people.


\section{Privacy in gaze estimation} \label{sec:eye_privacy}
\begin{figure}[b]
    \centering
    \includegraphics[width=\linewidth]{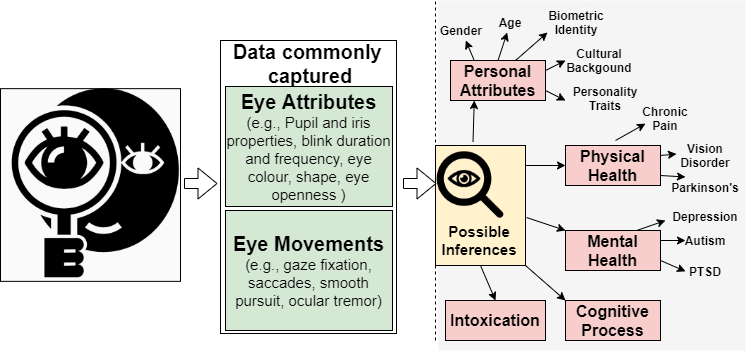}
    \caption{The possible privacy concerns related to gaze analysis framework~\cite{kroger2019does}. Please refer Sec.~\ref{sec:eye_privacy} for more details.}    
    \label{fig:eye_privacy}
    \vspace{-5mm}
\end{figure}

Due to the rapid progress over the past few years, gaze estimation technologies have become more reliable, cheap, compact and observe increasing use in many fields, such as gaming, marketing, driver safety, and healthcare. Consequently, these expanding uses of technology raise serious privacy concerns. Gaze patterns can reveal much more information than a user wishes and expects to give away. By portraying the sensitivity of gaze tracking data, this section provides a brief overview of privacy concerns and consequent implications of gaze estimation and eye-tracking. Fig.~\ref{fig:eye_privacy} shows the overview of the privacy concerns, including common data capturing scenarios with their possible implications. A recent analysis~\cite{kroger2019does} of the literature shows that eye-tracking data may implicitly contain information about a user's biometric identity~\cite{john2019eyeveil}, personal attributes (such as gender, age, ethnicity, personality traits, intoxication, emotional state, skills etc.)~\cite{erbilek2013age,moss2012eye,hoppe2018eye}, physical and mental health~\cite{harezlak2018application,alghowinem2013eye}. Few eye-tracking measures may even reveal underlying cognitive processes~\cite{eckstein2017beyond}. The widespread consideration of eye-tracking enhance the potential to improve our lives in many directions, but the technology can also pose a substantial threat to privacy. Thus, it is necessary to understand the sensitiveness of gaze data from a holistic perspective to prevent its misuse.

\section{Conclusion and Future Direction} \label{sec:conclusion_limit_future}
Gaze analysis is a technology in search of an application in several domains mainly in assistive technology and HCI. The wide applications of gaze related technology is growing rapidly. Thus, it opens a lot of research opportunity ahead of the community. Here, in this paper, we present an overall review of gaze analysis frameworks with different perspectives from different point of view. Beginning with the preliminaries of gaze modelling and eye movement, we further elaborate on challenges in this field, overview of gaze analysis framework and its possible applications in different domains. For eye analysis, mainly geometric and appearance properties are widely explored in prior works. Despite recent progress, the gaze analysis remains challenging due to eye head interplay, occlusion and other challenges mentioned in Sec.~\ref{sec:challenges}. Thus, there is a scope for future development in this respect. Moreover, all of the proposed datasets in this domain are collected in constraint environments. In order to overcome these limitations, the generative adversarial network based data generation approach has come into play. Due to several image quality-related issues, these datasets are not used for benchmarking. Automatic labelling of images based on accurate heuristic could be explored to reduce the data annotation burden greatly. Future directions for the eye and gaze trackers include:

\noindent \textbf{Gaze Analysis in Unconstrained Setup:} The most precise methods for gaze estimation is via intrusive sensors, IR camera and RGBD camera. The main drawback of these systems is that their performance degrades when used in real-world settings. In future, gaze estimation models should consider these situations. Although several current efforts in this direction employ techniques, yet further research is needed. Moreover, most of the current gaze estimation benchmark datasets require the proper geometric arrangement as well as user cooperation (e.g., CAVE, TabletGaze, MPII, Eyediap, ETH-XGaze etc). It would be an interesting direction to explore gaze estimation in a more flexible setting.


\noindent \textbf{Learning with Less Supervision:} With the surge in unsupervised, self-supervised, weakly supervised techniques in this domain, more exploration in this direction is required to eliminate the dependency on ground truth gaze label which could be error-prone due to data acquisition limitations. 

\noindent \textbf{Gaze Inference:} Apart from localizing the eye and determining gaze, the gaze patterns provides vital cues for encoding the cognitive and affective states of the concerned person. More exploration and cross-domain research could be another direction to encode visual perception.

\noindent \textbf{AR/VR:} Eye tracking has potential application in AR/VR including Foveated Rending (FR) and Attention Tunneling. The gaze based interaction require low latency gaze estimation. In these applications, the visual environment presents a high-quality image at the point where the user is looking while blurring the other peripheral region. The intuition is to reduce power consumption without compromising the perceptual quality  as well as user experience. However, eye movements are fast and involuntary action which restrict the use of this techniques (in FR) due to the subsequent delays in the eye-tracking pipelines. In order to address this issue, a new research direction i.e. future gaze trajectory prediction has been recently introduced~\cite{palmero2020openeds2020}. More exploration along this direction is highly desirable.

\noindent \textbf{Eye Model and Learning Based Hybrid Approaches:} Traditional geometrical eye model based and appearance guided learning based approaches have complimentary advantages. The geometrical eye model based methods does not require training data. Moreover, it has strong generalization capability but it is highly relied on relevant eye landmark localization performance. Accurate localization of eye landmarks is quite challenging in real world settings as the subject could have extreme headpose, occlusion, illumination and other environmental factors. On the other hand, the learning based approaches can encode eye appearance feature but it does not generalize well across different setups. Thus, a hybrid model which can take the advantage of both scenarios could be a possible research direction for gaze estimation and eye tracking domain.

\noindent \textbf{Multi-modal/Cross-modal Gaze Estimation:} Over the past decade, head gesture synthesis has become an interesting line of research. Prior works in this area have mainly used handcrafted audio features such as energy based features~\cite{ben2013articulatory}, MFCC (Mel Frequency Cepstral Coefficent)~\cite{ding2015head}, LPC (Linear Predictive Coding)~\cite{ding2015head} and filter bank~\cite{ding2015head,ding2015blstm} to generate realistic head gesture. The main challenge in this domain is audio data annotation for head motion synthesis which is a noisy and error prone process. Prior works approach this problem via multi-stream HMMs~\cite{ben2013articulatory}, MLP based regression modelling~\cite{ding2015head}, bi-LSTM~\cite{ding2015blstm} and Conditional Variational Autoencoder (CVAE)~\cite{greenwood2017predicting}. In vision domain, mainly visual stimuli is utilized for gaze estimation. As the audio signal is non-trivial for gaze estimation, yet, it has the potential to coarsely define the gaze direction~\cite{ghosh2022av}. Research along this direction have potential to estimate gaze in challenging situation where visual stimuli fails.

The techniques surveyed in this paper focus on gaze analysis from different perspective, however, these techniques can be useful for other computer vision and HCI tasks. Gaze analysis and its widespread applications is a unique and well-defined topic, which have already influenced recent technologies. Scholarly interest in gaze estimation is established in a large number of disciplines. It primarily originates from vision-related assistive technology which further propagates through other domains and attracts a lot of future research attention across various fields.




\ifCLASSOPTIONcaptionsoff
  \newpage
\fi

\bibliographystyle{IEEEtran}
\small{\bibliography{eye_gaze}}

%

\begin{IEEEbiography}[{\includegraphics[width=1in,height=1.25in,clip,keepaspectratio]{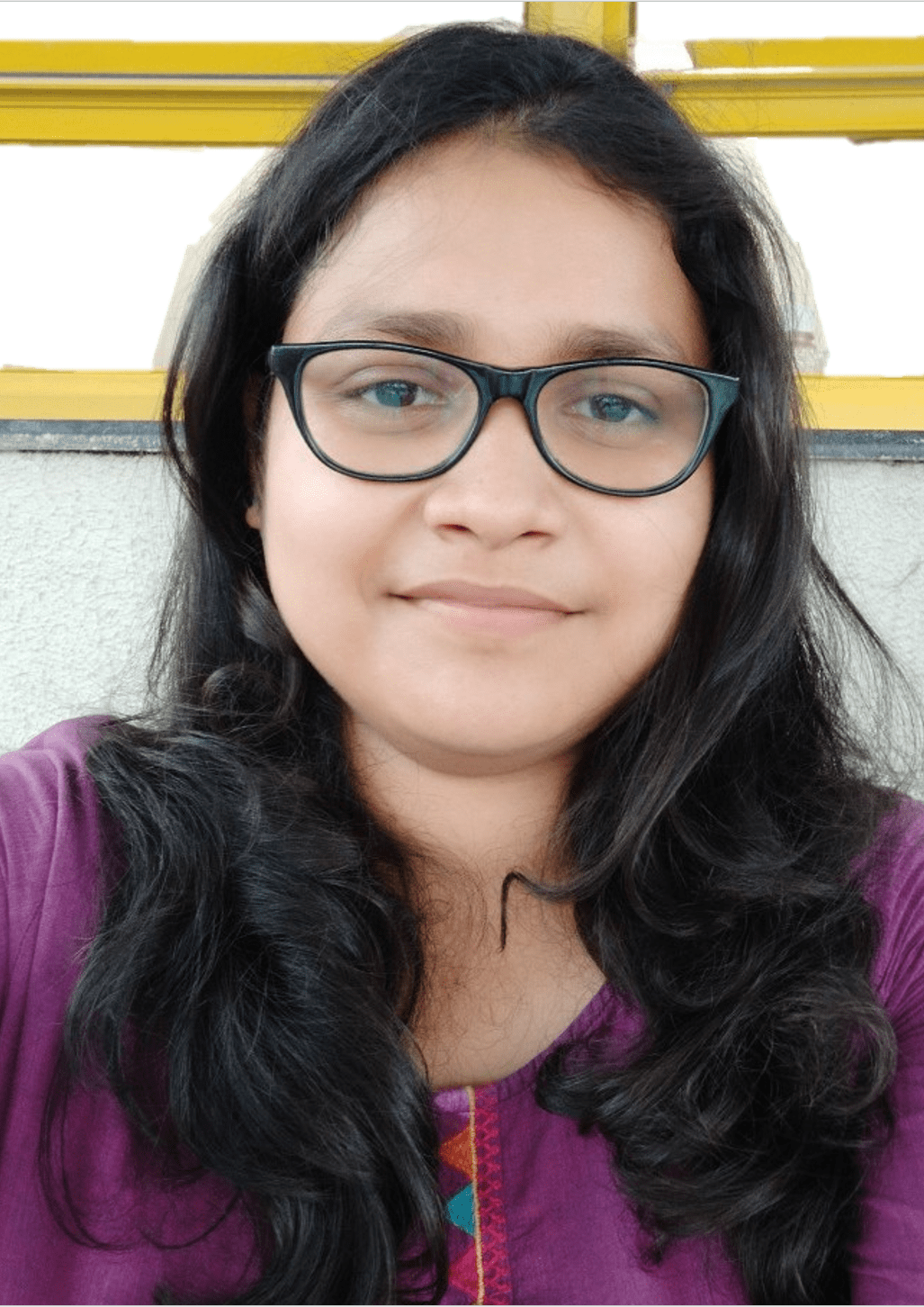}}]{Shreya Ghosh} is currently pursuing her PhD at Monash University, Australia. She received MS(R) degree in Computer Science and Engineering from the Indian Institute of Technology Ropar, India. She received the B.Tech. in CSE from the Govt. College of Engineering and Textile Technology (Serampore, India). Her research interests include affective computing, computer vision, Deep Learning. She is a student member of the IEEE.
\end{IEEEbiography}

\begin{IEEEbiography}[{\includegraphics[width=1in,height=1.25in,clip,keepaspectratio]{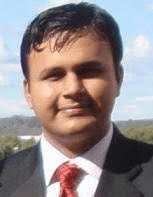}}]{Abhinav Dhall} is an Assistant Professor at Indian Institute of Technology Ropar and Adjunct Senior Lecturer at Monash University . He received PhD from the Australian National University in 2014. Followed by postdocs at the University of Waterloo and the University of Canberra. He was awarded the Best Doctoral Paper Award at ACM ICMR 2013, Best Student Paper Honourable mention at IEEE AFGR 2013 and Best Paper Nomination at IEEE ICME 2012. His research interests are in computer vision for Affective computing and Assistive Technology. He is a member of the IEEE and Associate Editor of IEEE Transactions on Affective Computing.
\end{IEEEbiography}

\begin{IEEEbiography}[{\includegraphics[width=1in,height=1.25in,clip,keepaspectratio]{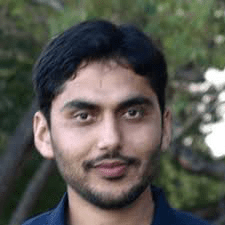}}]{Munawar Hayat} is currently a Senior Research Fellow with Monash University, Australia. He received his PhD from The University of Western Australia (UWA). His PhD thesis received multiple awards, including the Deans List Honorable Mention Award and the Robert Street Prize. After his PhD, he joined IBM Research as a postdoc and then moved to the University of Canberra as an Assistant Professor. He was a Senior Scientist at the Inception Institute of Artificial Intelligence, UAE. He has been awarded the ARC DECRA fellowship. His research interests are in computer vision, machine learning, deep learning, and affective computing. 
\end{IEEEbiography}
\begin{IEEEbiography}[{\includegraphics[width=1in,height=1.25in,clip,keepaspectratio]{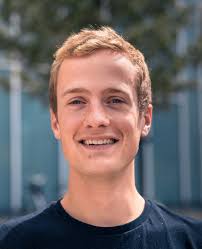}}]{Jarrod Knibbe} is currently with the University of Melbourne, Australia. He received his PhD from The University of Bristol in 2016. He completed a post-doc in human-centred computing at the University of Copenhagen. His research interests include interaction design and user experience, with a focus on body based user interfaces, electric muscle stimulation, and virtual reality. He has published over 25 papers at top venues in Human-Computer Interaction, including CHI, UIST, CSCW, and Ubicomp. 
\end{IEEEbiography}

\begin{IEEEbiography}[{\includegraphics[width=1in,height=1.25in,clip,keepaspectratio]{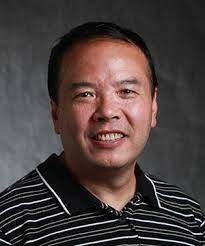}}]{Qiang Ji} is a Professor with the Department of Electrical, Computer, and Systems Engineering at Rensselaer Polytechnic Institute (RPI). He received his Ph.D degree in Electrical Engineering from the University of Washington. He was a program director at the National Science Foundation, where he managed NSF’s computer vision and machine learning programs. He also held teaching and research positions at University of Illinois at Urbana-Champaign, Carnegie Mellon University, and University of Nevada at Reno. His research interests are in computer vision, probabilistic machine learning, and their applications. He has published over 300 papers, received multiple awards for his work, serve as an editor for multiple international journals, and organize numerous international conferences/workshops. He is a fellow of the IEEE and the IAPR.
\end{IEEEbiography}







\end{document}